# Distributed Reasoning in a Peer-to-Peer Setting: Application to the Semantic Web

**Philippe Adjiman** ADJIMAN@LRI.FR
**Philippe Chatalic** CHATALIC@LRI.FR
**François Goasdoué** FG@LRI.FR
**Marie-Christine Rousset** MCR@LRI.FR
**Laurent Simon** SIMON@LRI.FR
*LRI-PCRI, Bâtiment 490*
*CNRS & Université Paris-Sud XI – INRIA Futurs*
*91405 Orsay Cedex, France*

## Abstract

In a peer-to-peer inference system, each peer can reason locally but can also solicit some of its acquaintances, which are peers sharing part of its vocabulary. In this paper, we consider peer-to-peer inference systems in which the local theory of each peer is a set of propositional clauses defined upon a local vocabulary. An important characteristic of peer-to-peer inference systems is that the global theory (the union of all peer theories) is not known (as opposed to partition-based reasoning systems). The main contribution of this paper is to provide the first consequence finding algorithm in a peer-to-peer setting: DeCA. It is anytime and computes consequences gradually from the solicited peer to peers that are more and more distant. We exhibit a sufficient condition on the acquaintance graph of the peer-to-peer inference system for guaranteeing the completeness of this algorithm. Another important contribution is to apply this general distributed reasoning setting to the setting of the Semantic Web through the Somewhere semantic peer-to-peer data management system. The last contribution of this paper is to provide an experimental analysis of the scalability of the peer-to-peer infrastructure that we propose, on large networks of 1000 peers.

## 1. Introduction

Recently peer-to-peer systems have received considerable attention because their underlying infrastructure is appropriate to scalable and flexible distributed applications over Internet. In a peer-to-peer system, there is no centralized control or hierarchical organization: each peer is equivalent in functionality and cooperates with other peers in order to solve a collective task. Peer-to-peer systems have evolved from simple keyword-based peer-to-peer file sharing systems like Napster (http://www.napster.com) and Gnutella (http://gnutella.wego.com) to semantic peer-to-peer data management systems like Edutella (Nejdl, Wolf, Qu, Decker, Sintek, & al., 2002) or Piazza (Halevy, Ives, Tatarinov, & Mork, 2003a), which handle semantic data description and support complex queries for data retrieval. In those systems, the complexity of answering queries is directly related to the expressivity of the formalism used to state the semantic mappings between peers schemas (Halevy, Ives, Suciu, & Tatarinov, 2003b).

In this paper, we are interested in peer-to-peer inference systems in which each peer can answer queries by reasoning from its local (propositional) theory but can also ask queries to some other peers with which it is semantically related by sharing part of its vocabulary. This framework encompasses several applications like peer-to-peer information integration systems or intelligent agents, in which each peer has its own knowledge (about its data or its expertise domain) and some partial knowledge about some other peers. In this setting, when solicited to perform a reasoning task, a peer, if it cannot solve completely that task locally, must be able to distribute appropriate reasoning subtasks among its acquainted peers. This leads to a step by step splitting of the initial task among the peers that are relevant to solve parts of it. The outputs of the different splitted tasks must then be recomposed to construct the outputs of the initial task.

We consider peer-to-peer inference systems in which the local theory of each peer is composed of a set of propositional clauses defined upon a set of propositional variables (called its local vocabulary). Each peer may share part of its vocabulary with some other peers. We investigate the reasoning task of finding consequences of a certain form (e.g., clauses involving only certain variables) for a given input formula expressed using the local vocabulary of a peer. Note that other reasoning tasks like finding implicants of a certain form for a given input formula can be equivalently reduced to the consequence finding task.

It is important to emphasize that the problem of distributed reasoning that we consider in this paper is quite different from the problem of reasoning over partitions obtained by decomposition of a theory (Dechter & Rish, 1994; Amir & McIlraith, 2000). In that problem, a centralized large theory is given and its structure is exploited to compute its best partitioning in order to optimize the use of a partition-based reasoning algorithm. In our problem, the whole theory (i.e., the union of all the local theories) is not known and the partition is imposed by the peer-to-peer architecture. As we will illustrate it on an example (Section 2), the algorithms based on transmitting clauses between partitions in the spirit of the work of Amir and McIlraith (2000), Dechter and Rish (1994) or del Val (1999) are not appropriate for our consequence finding problem. Our algorithm splits clauses if they share variables of several peers. Each piece of a splitted clause is then transmitted to the corresponding theory to find its consequences. The consequences that are found for each piece of splitted clause must then be recomposed to get the consequences of the clause that had been splitted.

The main contribution of this paper is to provide the first consequence finding algorithm in a peer-to-peer setting: DeCA. It is anytime and computes consequences gradually from the solicited peer to peers that are more and more distant. We exhibit a sufficient condition on the acquaintance graph of the peer-to-peer inference system for guaranteeing the completeness of this algorithm.

Another important contribution is to apply this general distributed reasoning setting to the setting of the Semantic Web through the Somewhere semantic peer-to-peer data management system. Somewhere is based on a simple data model made of taxonomies of atomic classes and mappings between classes of different taxonomies, which we think are the appropriate common semantic support needed for most of the future semantic web applications. The Somewhere data model can be encoded into propositional logic so that query answering in Somewhere can be equivalently reduced to distributed reasoning over logical propositional theories.

The last contribution of this paper is to provide an experimental analysis of the scalability of our approach, when deployed on large networks. So far, the scalability of a system like Piazza goes up to about 80 peers. Piazza uses a more expressive language than the one used in our approach, but its results rely on a wide range of optimizations (mappings composition, paths pruning Tatarinov & Halevy, 2004) made possible by the centralized storage of all the schemas and mappings in a global server. In contrast, we have stuck to a fully decentralized approach and performed our experiments on networks of 1000 peers.

An important point characterizing peer-to-peer systems is their dynamicity: peers can join or leave the system at any moment. Therefore, it is not feasible to bring all the information to a single server in order to reason with it locally using standard reasoning algorithms. Not only would it be costly to gather the data available through the system but it would be a useless task because of the changing peers connected to the network. The dynamicity of peer-to-peer inference systems imposes to revisit any reasoning problem in order to address it in a completely decentralized manner.

The paper is organized as follows. Section 2 defines formally the peer-to-peer inference problem that we address in this paper. In Section 3, we describe our distributed consequence finding algorithm and we state its properties. We describe Somewhere in Section 4. Section 5 reports the experimental study of the scalability of our peer-to-peer infrastructure. Related work is summarized in Section 6. We conclude with a short discussion in Section 7.

## 2. Consequence Finding in Peer-to-peer Inference Systems

A peer-to-peer inference system (P2PIS) is a network of peer theories. Each peer $P$ is a finite set of propositional formulas of a language $\mathcal{L}_P$. We consider the case where $\mathcal{L}_P$ is the language of clauses without duplicated literals that can be built from a finite set of propositional variables $\mathcal{V}_P$, called the *vocabulary* of $P$. Peers can be semantically related by sharing variables with other peers. A *shared variable* between two peers is in the intersection of the vocabularies of the two peers. We do not impose however that all the variables in common in the vocabularies of two peers are shared by the two peers: two peers may not be aware of all the variables that they have in common but only of some of them.

In a P2PIS, no peer has the knowledge of the global P2PIS theory. Each peer only knows its own local theory and the variables that it shares with some other peers of the P2PIS (its *acquaintances*). It does not necessarily knows *all* the variables that it has in common with other peers (including with its acquaintances). When a new peer joins a P2PIS it simply declares its acquaintances in the P2PIS, i.e., the peers it knows to be sharing variables with, and it declares the corresponding shared variables.

### 2.1 Syntax and Semantics

A P2PIS can be formalized using the notion of *acquaintance graph* and in the following we consider P2PIS and acquaintance graphs as equivalent.

**Definition 1 (Acquaintance graph)** *Let $\mathcal{P} = \{P_i\}_{i=1..n}$ be a collection of clausal theories on their respective vocabularies $\mathcal{V}_{P_i}$, let $\mathcal{V} = \cup_{i=1..n}\mathcal{V}_{P_i}$. An acquaintance graph over $\mathcal{V}$ is a graph $\Gamma = (\mathcal{P}, \text{ACQ})$ where $\mathcal{P}$ is the set of vertices and $\text{ACQ} \subseteq \mathcal{V} \times \mathcal{P} \times \mathcal{P}$ is a set of labelled edges such that for every $(v, P_i, P_j) \in \text{ACQ}$, $i \neq j$ and $v \in \mathcal{V}_{P_i} \cap \mathcal{V}_{P_j}$.*

A labelled edge $(v, P_i, P_j)$ expresses that peers $P_i$ and $P_j$ know each other to be sharing the variable $v$. For a peer $P$ and a literal $l$, $\text{ACQ}(l, P)$ denotes the set of peers sharing with $P$ the variable of $l$.

In contrast with other approaches (Ghidini & Serafini, 2000; Calvanese, De Giacomo, Lenzerini, & Rosati, 2004), we do not adopt an epistemic or modal semantics for interpreting a P2PIS but we interpret it with the standard semantics of propositional logic.

**Definition 2 (Semantics of a P2PIS)** *Let $\Gamma = (\mathcal{P}, \text{ACQ})$ be a P2PIS with $\mathcal{P} = \{P_i\}_{i=1..n}$,*

- *An* interpretation *$I$ of $\mathcal{P}$ is an assignement of the variables in $\bigcup_{i=1..n} P_i$ to true or false. In particular, a variable which is common to two theories $P_i$ and $P_j$ of a given P2PIS is interpreted by the same value in the two theories.*
- *$I$ is a* model of a clause *$c$ iff one of the literals of $c$ is evaluated to true in $I$.*
- *$I$ is a* model of a set of clauses *(i.e., a local theory, a union of a local theories, or a whole P2PIS) iff it is a model of all the clauses of the set.*
- *A P2PIS is* satisfiable *iff it has a model.*
- *The* consequence relation *for a P2PIS is the standard consequence relation $\models$: given a P2PIS $\mathcal{P}$, and a clause $c$, $\mathcal{P} \models c$ iff every model of $\mathcal{P}$ is a model of $c$.*

### 2.2 The Consequence Finding Problem

For each theory $P$, we consider a subset of *target variables* $\mathcal{TV}_P \subseteq \mathcal{V}_P$, supposed to represent the variables of interest for the application, (e.g., observable facts in a model-based diagnosis application, or classes storing data in an information integration application). The goal is, given a clause provided as an input to a given peer, to find all the possible consequences belonging to some *target language* of the input clause and the union of the peer theories.

The point is that the input clause only uses the vocabulary of the queried peer, but that its expected consequences may involve target variables of different peers. The target languages handled by our algorithm are defined in terms of target variables and require that a shared variable has the same target status in all the peers sharing it. It is worth noting that this requirement is a local property: the peers sharing variables with a given peer are its acquaintances and, by definition, they are its direct neighbours in the acquaintance graph.

**Definition 3 (Target Language)** *Let $\Gamma = (\mathcal{P}, \text{ACQ})$ be a P2PIS, and for every peer $P$, let $\mathcal{TV}_P$ be the set of its target variables such that if $(v, P_i, P_j) \in \text{ACQ}$ then $v \in \mathcal{TV}_{P_i}$ iff $v \in \mathcal{TV}_{P_j}$. For a set $SP$ of peers of $\mathcal{P}$, we define its* target language *$\mathcal{T}arget(SP)$ as the language of clauses (including the empty clause) involving only variables of $\bigcup_{P \in SP} \mathcal{TV}_P$.*

The reasoning problem that we are interested in is to compute logical consequences of an input clause given a P2PIS. It corresponds to the notion of proper prime implicates of a clause w.r.t. a clausal (distributed) theory, which if formally defined in Definition 4.

**Definition 4 (Proper prime implicate of a clause w.r.t. a clausal theory)** *Let $P$ be a clausal theory and $q$ be a clause. A clause $m$ is said to be:*

- *an* implicate *of $q$ w.r.t. $P$ iff $P \cup \{q\} \models m$.*
- *a* prime implicate *of $q$ w.r.t. $P$ iff $m$ is an implicate of $q$ w.r.t. $P$, and for any other clause $m'$ implicate of $q$ w.r.t. $P$, if $m' \models m$ then $m' \equiv m$.*
- *a* proper prime implicate *of $q$ w.r.t. $P$ iff it is a prime implicate of $q$ w.r.t. $P$ but $P \not\models m$.*

The problem of finding prime implicates from a new clause and a theory, a.k.a. $\Phi$-prime implicates, corresponds exactly to the problem of computing proper prime implicates of a clause w.r.t. a clausal theory. It has been extensively studied in the centralized case (see the work of Marquis, 2000, for a survey). Note that deciding whether a clause is a $\Phi$-prime implicate of a clausal theory is $BH_2$-complete (Marquis, 2000), i.e., both in $NP$ and $coNP$. The problem we address may be viewed as a further refinement, restricting to the computation of proper prime implicates of a given target language. It corresponds to $\langle L, \Phi\rangle$-prime implicates in the work of Marquis (2000) and has the same complexity.

**Definition 5 (The consequence finding problem in a P2PIS)** *Let $\Gamma = (\mathcal{P}, \text{ACQ})$ be a P2PIS, where $\mathcal{P} = \{P_i\}_{i=1..n}$ is a collection of clausal theories with respective target variables. The* consequence finding problem *in $\Gamma$ is, given a peer $P$, its acquaintances in $\Gamma$, and a clause $q \in \mathcal{L}_P$, to find the set of proper prime implicates of $q$ w.r.t. $\bigcup_{i=1..n} P_i$ which belong to $\mathcal{T}arget(\mathcal{P})$.*

From an algorithmic point of view, the consequence finding problem in a P2PIS is new and significantly different from the consequence finding problem in a single global theory. According to the semantics, in order to be complete, a peer-to-peer consequence finding algorithm must obtain the same results as any standard consequence finding algorithm applied to the union of the local theories, but without having it as a global input: it just has a partial and local input made of the theory of a single peer and of its acquaintances. The reasoning must be distributed appropriately among the different theories without a global view of the whole P2PIS. In a full peer-to-peer setting, such a consequence finding algorithm cannot be centralized (because it would mean that there is a super-peer controlling the reasoning). Therefore, we must design an algorithm running independently on each peer and possibly distributing part of reasoning that it controls to acquainted peers: no peer has control on the whole reasoning.

Among the possible consequences we distinguish *local consequences*, involving only target variables of the solicited peer, *remote consequences*, which involve target variables of a single peer distant from the solicited peer, and *combined consequences* which involve target variables of several peers.

### 2.3 Example

The following example illustrates the main characteristics of our message-based distributed algorithm running on each peer, which will be presented in detail in Section 3.

Let us consider 4 interacting peers. $P_1$ describes a tour operator. Its theory expresses that its current $Far$ destinations are either $Chile$ or $Kenya$. These far destinations are international destinations ($Int$) and expensive ($Exp$). The peer $P_2$ is only concerned with police regulations and expresses that a passport is required ($Pass$) for international destinations. $P_3$ focuses on sanitary conditions for travelers. It expresses that, in Kenya, yellow fever vaccination ($YellowFev$) is strongly recommended and that a strong protection against paludism should be taken ($Palu$) when accomodation occurs in $Lodges$. $P_4$ describes travel accommodation conditions : Lodge for Kenya and $Hotel$ for Chile. It also expresses that when anti-paludism protection is required, accommodations are equipped with appropriate anti-mosquito protections ($AntiM$). The respective theories of each peer are described on Figure 1 as the nodes of the acquaintance graph. Shared variables are mentioned as edge labels. Target variables are defined by : $\mathcal{TV}_{P_1} = \{Exp\}$, $\mathcal{TV}_{P_2} = \{Pass\}$, $\mathcal{TV}_{P_3} = \{Lodge, YellowFev, Palu\}$ and $\mathcal{TV}_{P_4} = \{Lodge, Hotel, Palu, AntiM\}$.

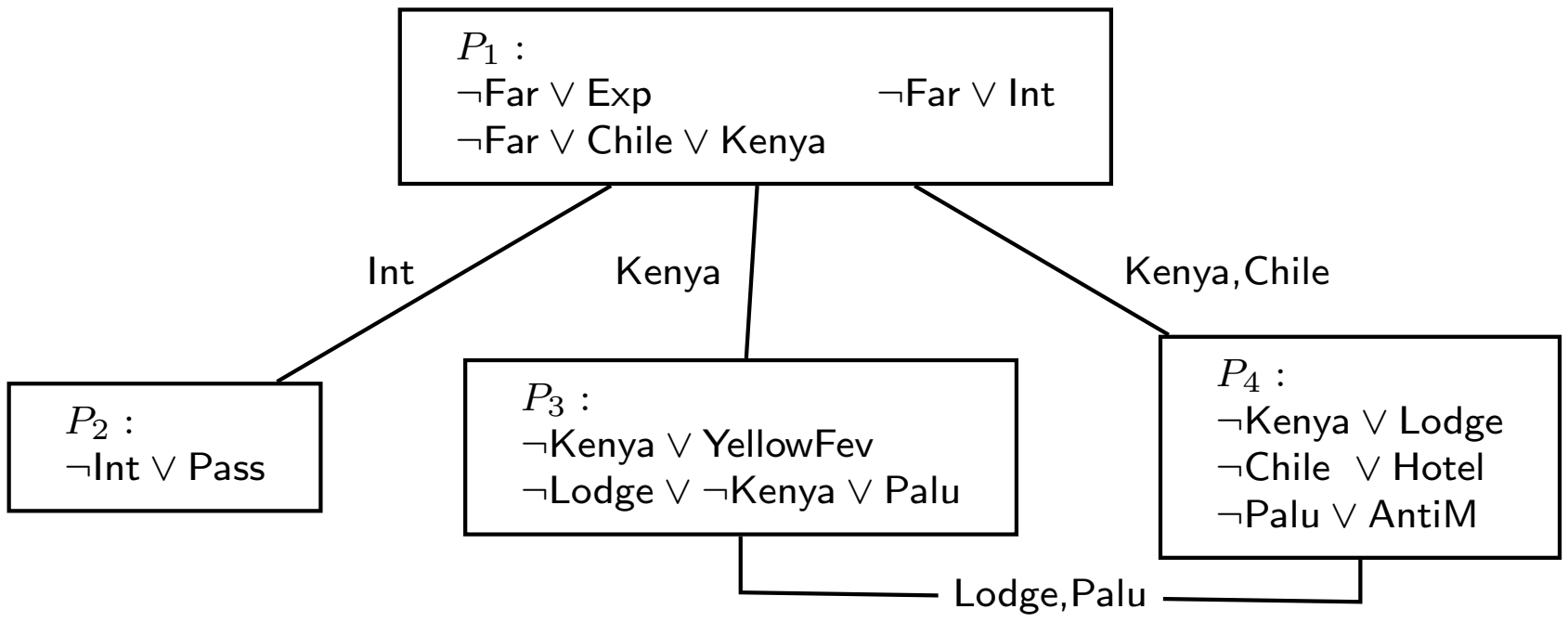

Figure 1: Acquaintance graph for the tour operator example

We now illustrate the behavior of the algorithm when the input clause Far is provided to peer $P_1$ by the user. Describing precisely the behavior of a distributed algorithm on a network of peers is not easy. In the following we present the propagation of the reasoning as a tree structure, the nodes of which correspond to peers and the branches of which materialize the different reasoning paths induced by the initial input clause. Edges are labelled on the left side by literals which are propagated along paths and/or on the right side by consequences that are transmitted back. A downward arrow on an edge indicates the step during which a literal is propagated from one peer to its neighbor. For instance, the initial step can be represented here by the following tree :

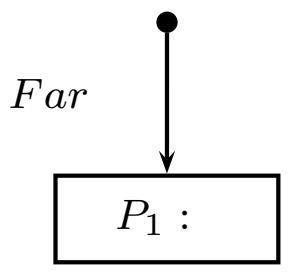

Local consequences of a literal propagated on a peer are then explicited within the peer node. Target literals are outlined using a grey background, as well as transmitted back consequences. Vertical arrows preceding consequences distinguish the last returned consequences from earlier ones. Although the successive trees presented here have increasing depth, as if all reasoning paths were explored synchronously and in parallel, the reader should keep in mind that all messages are exchanged in an asynchronous way and that the order in which consequents are produced cannot be predicted.

In our example, consequences of Far derived by local reasoning on $P_1$ are Exp, Int and Chile $\vee$ Kenya. Since Exp is in $\mathcal{T}arget(P_1)$ it is a local consequence of Far. Int is not a target literal but is shared with $P_2$, it is therefore transmitted to $P_2$. The clause Chile $\vee$ Kenya is also made of shared variables. Such clauses are processed by our algorithm using a split/recombination approach. Each shared literal is processed independently, and transmitted to its appropriate neighbors. Each literal is associated with some queue data structure, where transmitted back consequences are stored. As soon as at least one consequent has been obtained for each literal, the respective queued consequents of each literal are recombined, to produce consequences of the initial clause. This recombination process continues, as new consequences for a literal are produced. Note that since each literal is processed asynchronously, the order in which the recombined consequences are produced is unpredictable. Here, the component Chile is transmitted to $P_4$ and Kenya is transmitted to $P_3$ and $P_4$. Let us note that the peer $P_4$ appears two times in the tree, because two different literals are propagated on this peer, which induces two different reasoning paths.

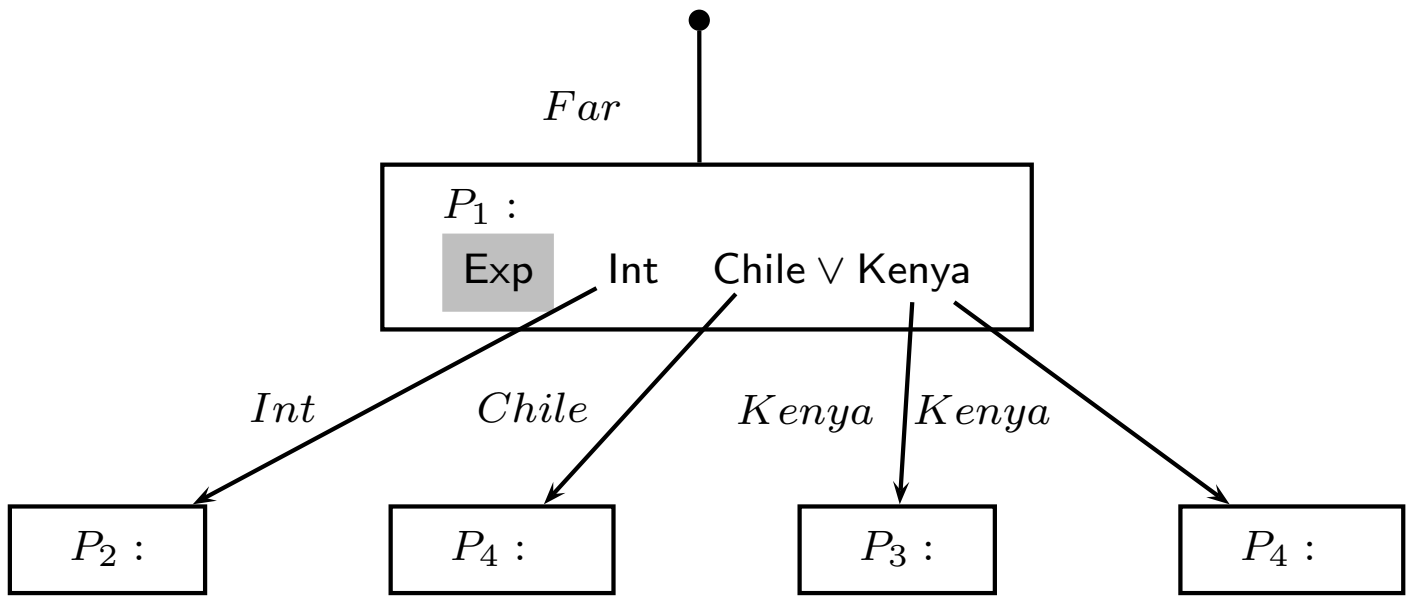

While Exp is transmitted back to the user as a first (*local*) consequence of Far.

- The propagation of Int on $P_2$ produces the clause Pass, which is in $\mathcal{T}arget(P_2)$ but is not shared and therefore, cannot be further propagated.

- The clause Chile, when transmitted to $P_4$, produces Hotel which is in $\mathcal{T}arget(P_4)$ but is not shared and cannot be further propagated.

- When transmitted to $P_3$, the clause Kenya produces YellowFev as well as the clause $\neg$Lodge $\vee$ Palu. The three variables are in $\mathcal{T}arget(P_3)$. Lodge and Palu are also shared variables and therefore, after splitting of the second clause, their corresponding literals are transmitted (independently) to $P_4$.

- When transmitted to $P_4$, Kenya produces Lodge, which is in $\mathcal{T}arget(P_4)$ and is also shared and therefore further transmitted to $P_3$.

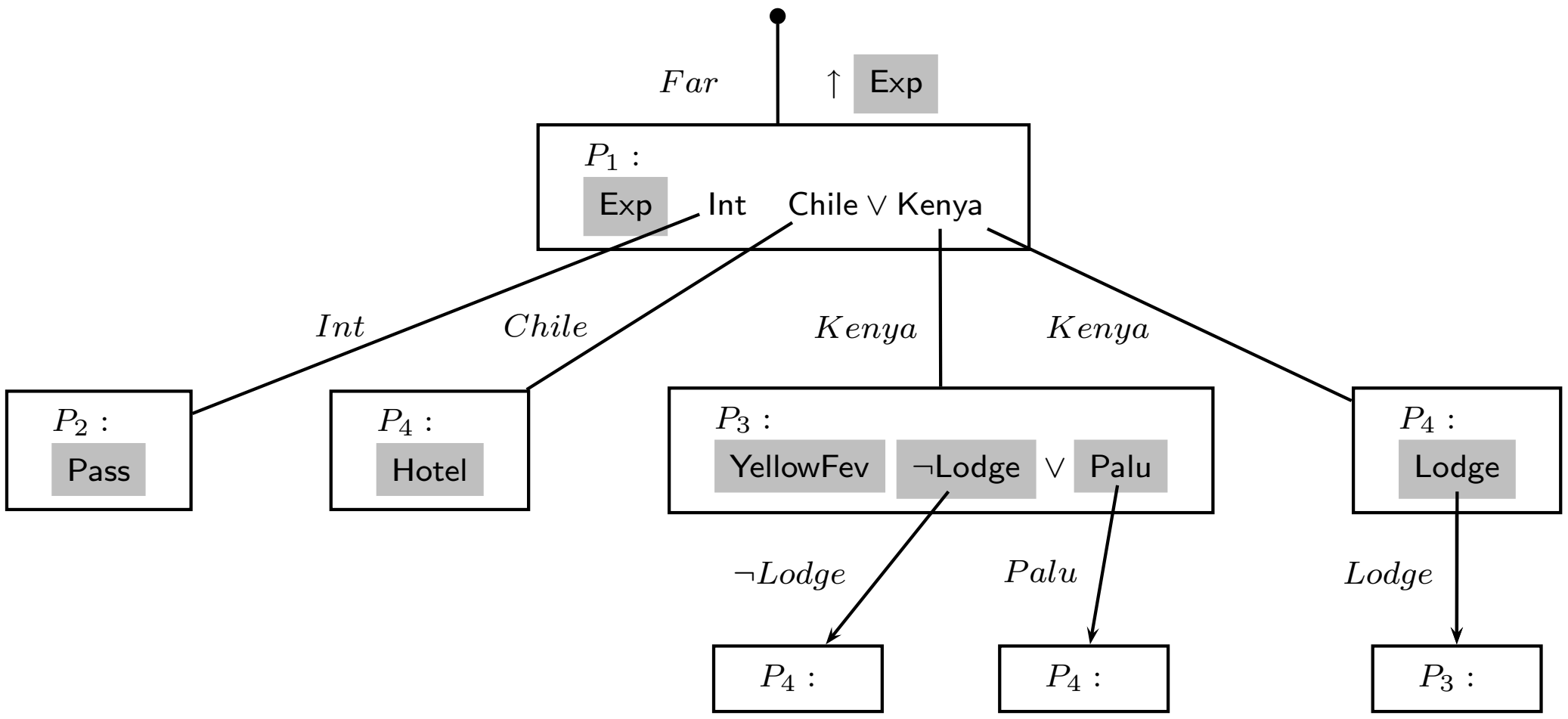

- The clause Pass, produced on $P_2$, is transmitted back to $P_1$ as a consequence of Int and then to the user as a *remote consequence* of Far.

- The clause Hotel, produced on $P_4$, is transmitted back to $P_1$ where it is queued as a consequent of Chile, since it has to be combined with consequences of Kenya.

- The two local consequences of Kenya obtained on $P_3$ contain only target variables. They are transmitted back to $P_1$ and queued there. They may now be combined with Hotel to produce two new *combined consequences* of Far : Hotel ∨ YellowFev and Hotel ∨ ¬Lodge ∨ Palu, which are transmitted back to the user.

- Similarly on $P_4$, Lodge is a local target consequent of Kenya, that is transmitted back to $P_1$ as a consequent of Kenya, where it is combined with Hotel to produce a new consequence of Far that, in turn, is transmitted back to the user.

Simultaneously, the reasoning further propagates in the network of peers. The propagation of ¬Lodge and Palu on $P_4$ respectively produces ¬Kenya, which is not a target literal but is shared and thus further propagated on $P_1$, as well as AntiM, which is a target literal, but not shared. We do not detail here the propagation of Lodge in the right most branch of the reasoning tree.

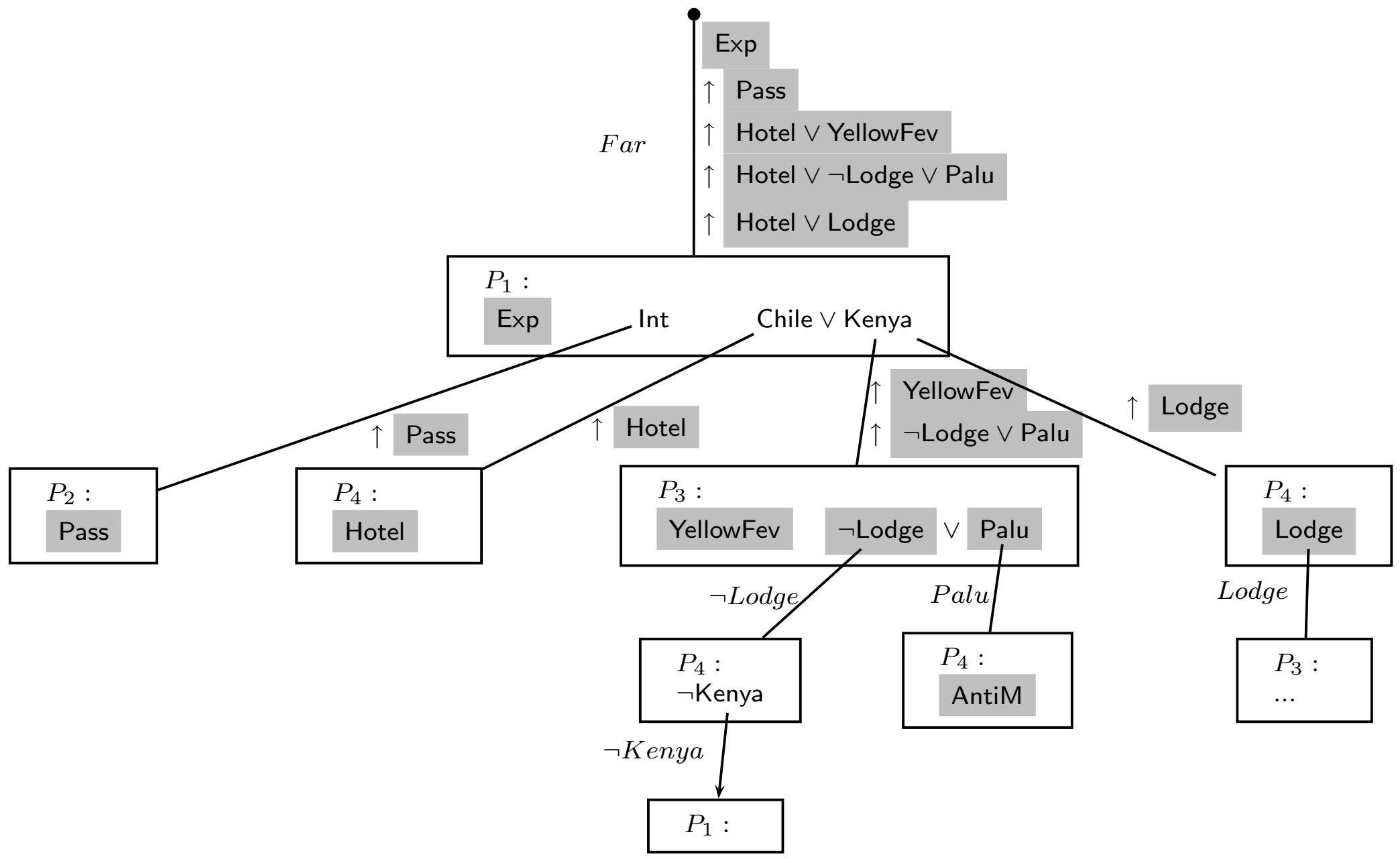

Note on the deepest node that $P_1$ is here asked to produce the implicates of ¬Kenya, while the complementary literal Kenya is still under process. We will see in Section 3 that such situations are handled in our algorithm by mean of histories keeping track of the reasoning branches ending up to each transmitted literal. When a same history contains two complementary literals, the corresponding reasoning branch is closed and the empty clause □ is returned as a consequence of the literals in the history.

In our example, the consequence produced by $P_1$ for ¬Kenya is thus □, which is sent back to $P_4$ and then to $P_3$. After combination on $P_3$ with Palu we thus obtain Palu as a new consequent of Kenya, which subsumes the previously obtained ¬Lodge ∨ Palu. When transmitted back to $P_1$ and combined with Hotel we obtain Hotel ∨ Palu which subsumes the previously obtained consequent Hotel ∨ ¬Lodge ∨ Palu. Since AntiM is not a shared variable it is the only consequent of Palu on $P_4$. When transmitted back to $P_3$ for combination with □, we thus obtain AntiM which, in turn, is returned to $P_1$ for combination with Hotel, thus giving Hotel ∨ AntiM as a new consequent of Far.

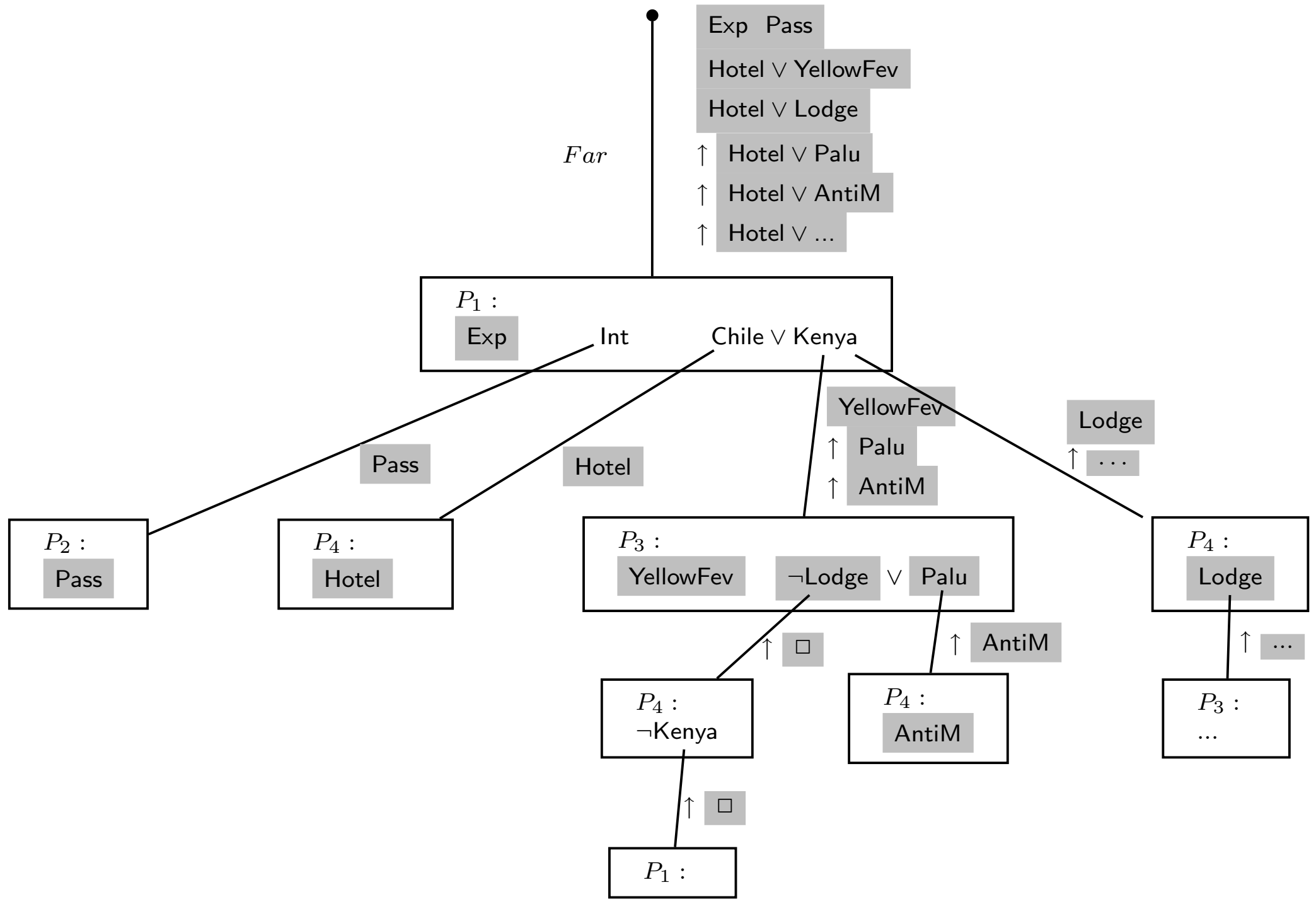

We have not detailed the production of consequences of Lodge on $P_3$ in the right most branch. The reader could check in a similar way that it also produces AntiM (which has already been produced through $P_3/P_4$ in the third branch). Eventually, the whole set of consequences of Far is {Exp, Pass, Hotel∨Lodge, Hotel∨Palu, Hotel∨AntiM, Hotel∨YellowFev}. Among those consequences, it is important to note that some of them (e.g., Hotel∨YellowFev) involve target variables from different peers. Such implicates could not be obtained by partition-based algorithms like those in (Amir & McIlraith, 2000). This is made possible thanks to the split/recombination strategy of our algorithm.

## 3. Distributed Consequence Finding Algorithm

The message passing distributed algorithm that we have implemented is described in Section 3.2. We show that it terminates and that it computes the same results as the recursive algorithm described in Section 3.1. We exhibit a property of the acquaintance graph that guarantees the completeness of this recursive algorithm, and therefore of the message passing distributed algorithm (since both algorithms compute the same results).
For both algorithms, we will use the following notations :

- for a literal $q$, $Resolvent(q, P)$ denotes the set of clauses obtained by resolution from the set $P \cup \{q\}$ but not from $P$ alone. We call such clauses *proper resolvents of q w.r.t. P,*
- *for a literal q,* $\bar{q}$ *denotes its complementary literal,*
- *for a clause c of a peer P,* $S(c)$ *(resp.* $L(c)$*) denotes the disjunction of literals of c whose variables are shared (resp. not shared) with some acquaintance of P. The* *condition $S(c) = \square$ thus expresses that $c$ does not contain any variable shared with an acquaintance of $P$,*

- *a history hist is a sequence of triples $(l, P, c)$ (where $l$ is a literal, $P$ a peer, and $c$ a clause). An history $[(l_n, P_n, c_n), \ldots, (l_1, P_1, c_1), (l_0, P_0, c_0)]$ represents a branch of reasoning initiated by the propagation of the literal $l_0$ within the peer $P_0$, which either contains the clause $\neg l_0 \vee c_0$ (in that case $c_0$ may have been splitted into its different literals among which $l_1$ is propagated in $P_1$), or not (in that case $l_0 = c_0$ and $l_0$ is propagated into $P_1$, and thus $l_0 = l_1$): for every $i \in [0..n-1]$, $c_i$ is a consequence of $l_i$ and $P_i$, and $l_{i+1}$ is a literal of $c_i$, which is propagated in $P_{i+1}$,*

- *$\otimes$ is the distribution operator on sets of clauses: $S_1 \otimes \cdots \otimes S_n = \{c_1 \vee \cdots \vee c_n \,|\, c_1 \in S_1, \ldots, c_n \in S_n\}$. If $L = \{l_1, \ldots, l_p\}$, we will use the notation $\otimes_{l \in L} S_l$ to denote $S_{l_1} \otimes \cdots \otimes S_{l_p}$.*

### 3.1 Recursive Consequence Finding Algorithm

Let $\Gamma = (\mathcal{P}, \textsc{acq})$ be a P2PIS, $P$ one of its peers, and $q$ a literal whose variable belongs to the vocabulary of $P$. $RCF(q, P)$ computes implicates of the literal $q$ w.r.t. $\mathcal{P}$, starting with the computation of local consequences of $q$, i.e., implicates of $q$ w.r.t. $P$, and then recursively following the acquaintances of the visited peers. To ensure termination, it is necessary to keep track of the literals already processed by peers. This is done by the recursive algorithm $RCFH(q, SP, hist)$, where $hist$ is the history of the reasoning branch ending up to the propagation of the literal $q$ in $SP$, which is the set of acquaintances of the last peer added to the history.

**Algorithm 1:** Recursive consequence finding algorithm

$RCF(q, P)$
**return** $RCFH(q, \{P\}, \emptyset)$

$RCFH(q, SP, hist)$
**if** there exists $P \in SP$ s.t. $q \in P$ or if for every $P \in SP$, $(q, P, _) \in hist$ **return** $\emptyset$
**else if** $(\bar{q}, _, _) \in hist$ **return** $\{\square\}$
**else** for every $P \in SP$ $\textsc{local}(P) \leftarrow \{q\} \cup Resolvent(q, P)$
**if** there exists $P \in SP$ s.t. $\square \in \textsc{local}(P)$ **return** $\{\square\}$
**else** for every $P \in SP$ $\textsc{local}(P) \leftarrow \{c \in \textsc{local}(P) | L(c) \in \mathcal{T}arget(P)\}$
**if** for every $P \in SP$ and for every $c \in \textsc{local}(P)$, $S(c) = \square$, **return** $\bigcup_{P \in SP} \textsc{local}(P)$
**else**
  $\textsc{result} \leftarrow \bigcup_{P \in SP} \{c \in \textsc{local}(P) | S(c) \in \mathcal{T}arget(P)\}$
  **foreach** $P \in SP$ and $c \in \textsc{local}(P)$ s.t. $S(c) \neq \square$
    **if** $\neg q \vee c \in P$, $P \leftarrow P \backslash \{\neg q \vee c\}$
    **foreach** literal $l \in S(c)$
      $\textsc{answer}(l) \leftarrow RCFH(l, \textsc{acq}(l, P), [(q, P, c) | hist])$
    $\textsc{disjcomb} \leftarrow (\otimes_{l \in S(c)} \textsc{answer}(l)) \otimes \{L(c)\}$
    $\textsc{result} \leftarrow \textsc{result} \cup \textsc{disjcomb}$
  **return** $\textsc{result}$

We now establish properties of this algorithm. Theorem 1 states that the algorithm is guaranteed to terminate and that it is sound. Theorem 2 exhibits a condition on the acquaintance graph for the algorithm to be complete. For the properties of soundness and completeness, we consider that the topology and the content of the P2PIS do not change while the algorithm is running. Therefore, those properties have the following meaning for a P2PIS: the algorithm is sound (respectively, complete) iff for every P2PIS, the results returned by $RCF(q, P)$, where $P$ is any peer of the P2PIS and $q$ any literal whose variable belongs to the vocabulary of $P$, are implicates (respectively, include all the proper prime implicates) of $q$ w.r.t. the union of all the peers in the P2PIS, *if there is no change in the P2PIS while the algorithm is running.*

The sufficient condition exhibited in Theorem 2 for the completeness of the algorithm is a global property of the acquaintance graph: any two peers having a variable in common must be either acquainted (i.e., must share that variable) or must be related by a path of acquaintances sharing that variable. First, it is important to emphasize that even if that property is global, it has not to be checked before running the algorithm. If it is not verified, the algorithm remains sound but not complete. Second, it is worth noticing that the modeling/encoding of applications into our general peer-to-peer propositional reasoning setting may result in acquaintance graphs satisfying that global property by construction. In particular, as it will be shown in Section 4 (Proposition 2), it is the case for the propositional encoding of the Semantic Web applications that we deal with in Somewhere.

**Theorem 1** *Let $P$ be a peer of a P2PIS and $q$ a literal belonging to the vocabulary of $P$. $RCF(q, P)$ is sound and terminates.*

Proof: **Soundness:** We need to prove that every result returned by $RCF(q, P)$ belongs to the target language and is an implicate of $q$ w.r.t. $\mathcal{P}$, where $\mathcal{P}$ is the union of all the peers in the P2PIS. For doing so, we prove by induction on the number $rc$ of recursive calls of $RCFH(q, SP, hist)$ that every result returned by $RCFH(q, SP, hist)$ (where the history $hist$, if not empty, is of the form $[(l_n, P_n, c_n), \dots, (l_0, P_0, c_0)]$) is an implicate of $q$ w.r.t. $\mathcal{P} \cup \{l_n, \dots, l_0\}$ which belongs to the target language.

- $rc = 0$: either one of the conditions of Line (1), Line (2), Line (4) or Line (6) is satisfied.

  - If the condition in Line (1) is satisfied, the algorithm returns an empty result.
  - If either there exists a peer $P$ such that $(\bar{q}, _, _) \in hist$ or $\square \in$ local$(P)$: in both cases, $\square$ is returned by the algorithm (in respectively Line (2) and Line (4)) and it is indeed an implicate of $q$ w.r.t. $\mathcal{P} \cup \{l_n, \dots, l_0\}$ belonging to the target language.

  - Let $r$ be a result returned by the algorithm at Line (6): there exists $P \in SP$ such that $r \in$ local$(P)$, and it is obvioulsy an implicate of $q$ w.r.t. $\mathcal{P} \cup \{l_n, \dots, l_0\}$ (as $q$ or a resolvent of $q$ and of a clause of $P$), and it belongs to the target language.

- Suppose the induction hypothesis true for $rc \leq p$, and let $SP$ be a set of peers of a P2PIS and $q$ a literal (belonging to the vocabulary of all the peers of $SP$) such that $RCFH(q, SP, hist)$ requires $p + 1$ recursive calls to terminate. Let $r$ be a result returned by $RCFH(q, SP, hist)$.

- If $r \in \text{LOCAL}(P)$ for a $P \in SP$ and is such that $S(r) = \Box$ or $S(r) \in \mathcal{T}arget(P)$, it is obviously an implicate of $q$ w.r.t. $\mathcal{P} \cup \{l_n, \dots, l_0\}$ belonging to the target language.

- If $r \notin \text{LOCAL}(P)$ for any $P \in SP$, it is obtained at Line (13): there exist $P \in SP$ and a clause $c$ of $P$ of the form $S(c) \vee L(c)$ such that $S(c) = ll_1 \vee \dots \vee ll_k$ and $r = r_1 \vee \dots \vee r_k \vee L(c)$, where every $r_i$ is a result returned by $RCFH(ll_i, \text{ACQ}(ll_i, P), [(q, P, c)|hist])$ (Line (12)). According to the induction hypothesis (the number of recursive calls of $RCFH(ll_i, \text{ACQ}(ll_i, P), [(q, P, c)|hist])$ for every $ll_i$ is less than or equal to $p$), every $r_i$ belongs to the target language and is an implicate of $ll_i$ w.r.t. $\mathcal{P} \backslash \{\neg q \vee c\} \cup \{q, l_n, \dots, l_0\}$, or, equivalently, an implicate of $q$ w.r.t. $\mathcal{P} \backslash \{\neg q \vee c\} \cup \{ll_i, l_n, \dots, l_0\}$.

Therefore, $r_1 \vee \dots \vee r_k$ belongs to the target language and is an implicate of $q$ w.r.t. $\mathcal{P} \backslash \{\neg q \vee c\} \cup \{S(c), l_n, \dots, l_0\}$. Since $L(c)$ belongs to the target language and $c = S(c) \vee L(c)$, $r$ (i.e, $r_1 \vee \dots \vee r_k \vee L(c)$) belongs to the target language, and is an implicate of $q$ w.r.t. $\mathcal{P} \backslash \{\neg q \vee c\} \cup \{c, l_n, \dots, l_0\}$, and a fortiori of $q$ w.r.t. $\mathcal{P} \cup \{c, l_n, \dots, l_0\}$ Since $c \in \text{LOCAL}(P)$, $c$ is an implicate of $q$ w.r.t. $\mathcal{P}$, and therefore $r$ is an implicate of $q$ w.r.t. $\mathcal{P} \cup \{l_n, \dots, l_0\}$.

**Termination:** at each recursive call, a *new* triple $(sl, P, c)$ is added to the history. If the algorithm did not terminate, the history would be infinite, which is not possible since the number of peers, literals and clauses within a P2PIS is finite.

□

The following theorem exhibits a sufficient condition for the algorithm to be complete.

**Theorem 2** *Let $\Gamma = (\mathcal{P}, \text{ACQ})$ be a P2PIS. If for every $P$, $P'$ and $v \in \mathcal{V}_P \cap \mathcal{V}_{P'}$ there exists a path between $P$ and $P'$ in $\Gamma$, all edges of which are labelled with $v$, then for every literal $q \in \mathcal{L}_P$, $RCF(q, P)$ computes all the proper prime implicates of $q$ w.r.t. $\mathcal{P}$ which belong to $\mathcal{T}arget(\mathcal{P})$.*

Proof: In fact, we prove that $RCF(q, P)$ computes at least all the *prime* proper resolvents of $q$ w.r.t. $P$, i.e., the elements of $Resolvent(q, P)$ that are not strictly subsumed by other elements of $Resolvent(q, P)$.

We first show that proper prime implicates of $q$ w.r.t. $P$ are prime proper resolvents of $q$ w.r.t. $P$. Let $m$ be a proper prime implicate of $q$ w.r.t. $P$. By definition $P \cup \{q\} \models m$ and $P \not\models m$. By completeness of resolution w.r.t. prime implicates, $m$ can be obtained by resolution from the set $P \cup \{q\}$ but not from $P$ alone, i.e., it is a proper resolvent of $q$ w.r.t. $P$. Let us suppose that $m$ is strictly subsumed by another element $m'$ of $Resolvent(q, P)$. This means that $P \cup \{q\} \models m' \models m$ and $m \not\equiv m'$, which contradicts that $m$ is a prime implicate of $q$ w.r.t. $P$.

Now we prove by induction on the maximum number $rc$ of recursive calls involving a same literal when triggering $RCFH(q, SP, hist)$ that $RCFH(q, SP, hist)$ computes all the proper prime implicates belonging to the target language of $q$ w.r.t. $\mathcal{P}(hist)$, where $\mathcal{P}(hist)$ is obtained from $\mathcal{P}$ by replacing each $\neg l_i \vee c_i$ by $l_i$ if $l_i \neq c_i$. Thus:

$\mathcal{P}(hist) = \mathcal{P}$ if $hist$ is empty,

otherwise:

$\mathcal{P}(hist) = \mathcal{P} \backslash \{\neg l_i \vee c_i | (l_i, P_i, c_i) \in hist \text{ s.t. } l_i \neq c_i\} \cup \{l_i | (l_i, P_i, c_i) \in hist \text{ s.t. } l_i \neq c_i\}$

If the history *hist* is not empty, it is of the form $[(l_n, P_n, c_n), \ldots, (l_0, P_0, c_0)]$. According to the algorithm, when $RCFH(q, SP, hist)$ is triggered, there have been at least $n+1$ previous calls of the algorithm $RCFH$: $RCFH(l_0, SP_0, \emptyset)$ and $RCFH(l_i, SP_i, hist_i)$ for $i \in [1..n]$, where $hist_i = [(l_{i-1}, P_i, c_{i-1}) \ldots, (l_0, P_0, c_0)]$), and $P_i \in SP_i$ for every $i \in [0..n]$. Since the P2PIS can be cyclic, it may be the case that when we call $RCFH(q, SP, hist)$, $(q, P, q) \in hist$. If that is the case, there have been previous calls of $RCFH$ involving $q$, i.e., of the form $RCFH(q, SP_i, hist_i)$.

- $rc = 0$: either one of the conditions of Line (1), Line (2), Line (4) or Line (6) is satisfied.

  - If the first condition is satisfied, since $rc = 0$, it cannot be the case that for every $P \in SP$, $(q, P, _) \in hist$, and therefore there exists $P \in SP$ such that $q \in P$: in this case, there is no *proper* prime implicate of $q$ w.r.t. $\mathcal{P}(hist)$, because $q \in \mathcal{P}(hist)$ and all the prime implicates of $q$ w.r.t. a theory containing $q$ are consequences of that theory.

  - If either $(\bar{q}, _, _) \in hist$ or $\square \in \text{LOCAL}(P)$ for a given peer $P$ of $SP$: in both cases, $\square$ is the only prime implicate of $q$ w.r.t. $\mathcal{P}(hist)$ and therefore, if it is a *proper* prime implicate, it is the only one too. It is returned by the algorithm (respectively Line (2) and Line (4)).

  - If for every $P \in SP$, every resolvent of $q$ w.r.t. $P$ has no shared variable with any acquaintance of $P$: if $\mathcal{P}$ satisfies the property stated in the theorem, this means that every prime implicate of $q$ w.r.t. $P$ has no variable in common with any other theory of $\mathcal{P}$. According to Lemma 1, the set of proper resolvents of $q$ w.r.t. $\mathcal{P}(hist)$ is included in $\bigcup_{P \in SP} \text{LOCAL}(P)$, and thus in particular every proper prime implicate of $q$ w.r.t. $\mathcal{P}(hist)$, which is in the target language, is returned by the algorithm (Line(6)).

- Suppose the induction hypothesis true for $rc \leq p$, and let $SP$ be a set of peers of a P2PIS satisfying the property stated in the theorem, such that $RCFH(q, SP, hist)$ requires atmost $p+1$ recursive calls involving $q$. Since there is at least one recursive call, the condition of Line (1) is not satisfied. Let $m$ be in the target language and a proper prime implicate of $q$ w.r.t. $\mathcal{P}(hist)$, where $\mathcal{P}(hist) = \mathcal{P} \backslash \{\neg l_i \vee c_i | (l_i, P_i, c_i) \in hist \text{ s.t. } l_i \neq c_i\} \cup \{l_i | (l_i, P_i, c_i) \in hist \text{ s.t. } l_i \neq c_i\}$. Let us show that $m$ belongs to the result returned by $RCFH(q, SP, hist)$.

  - If $m$ is a proper resolvent of $q$ w.r.t. a given $P$ of $SP$, then $m \in \text{LOCAL}(P)$ and is returned by the algorithm since it is in the target language.

  -If $m$ is not a proper resolvent of $q$ w.r.t. a given $P$ of $SP$, then, according to Lemma 1, either (i) $q$ has its variable in common with clauses in $\mathcal{P}(hist) \backslash \bigcup_{P \in SP} P$, or (ii) there exists a clause $\neg q \vee c$ in $\bigcup_{P \in SP} P$ such that $c$ has variables in common with $\mathcal{P}(hist) \backslash \bigcup_{P \in SP} P$ and $m$ is a proper resolvent of $c$ w.r.t. $\mathcal{P}(hist) \backslash \{\neg q \vee c\} \cup \{q\}$. In addition, it is a *prime* proper resolvent of $c$ w.r.t. $\mathcal{P}(hist) \backslash \{\neg q \vee c\} \cup \{q\}$. Let us suppose that this is not the case. Then there exists some clause $m' \in Resolvent(c, \mathcal{P}(hist) \backslash \{\neg q \vee c\} \cup \{q\})$, such that $m' \models m$ and $m' \not\equiv m$. By soundness, $\mathcal{P}(hist) \backslash \{\neg q \vee c\} \cup \{q\} \cup \{c\} \models m'$. Since $\mathcal{P}(hist) \backslash \{\neg q \vee c\} \cup \{q\} \cup \{c\} \equiv \mathcal{P}(hist) \cup \{q\}$, $\mathcal{P}(hist) \cup \{q\} \models m'$ with $m' \models m$ and $m' \not\equiv m$, which contradicts that $m$ is a prime implicate of $q$ w.r.t. $\mathcal{P}(hist)$.

(i) In the first case, according to the property stated in the theorem, the variable of $q$ is shared with other peers of the P2PIS than those in $SP$, and therefore $q$ is involved in an iteration of the loop of Line (9). According to the induction hypothesis (the number of recursive calls to obtain ANSWER($q$) in Line (12) is less than or equal to $p$) ANSWER($q$) includes the set of proper prime resolvents of $q$ w.r.t. $\mathcal{P}(hist')$, which are in the target language, where $hist' = [(q, P, q)|hist]$, and thus $\mathcal{P}(hist') = \mathcal{P}(hist)$. Therefore, ANSWER($q$) includes the set of proper prime resolvents of $q$ w.r.t. $\mathcal{P}(hist)$, in particular $m$.

(ii) In the second case, according to the property stated in the theorem, $c$ shares variables with other peers of the P2PIS than those in $SP$. In addition, since $m$ is in the target language, the local variables of $c$ are target variables. Therefore $c$ is involved in an iteration of the loop of Line (9). According to the induction hypothesis (the number of recursive calls to obtain ANSWER($l$) in Line (12) is less than or equal to $p$), for every $l \in S(c)$, ANSWER($l$) includes the set of all proper prime resolvents of $l$ w.r.t. $\mathcal{P}(hist')$ and thus in particular, the set of all proper prime implicates of $l$ w.r.t. $\mathcal{P}(hist')$ which are in the target language. Since $hist' = [(q, P, c)|hist]$ with $q \neq c$ (because there is no duplicate literals in the clauses that we consider), $\mathcal{P}(hist') = \mathcal{P}(hist) \backslash \{\neg q \vee c\} \cup \{q\}$. We can apply Lemma 2 to infer that $DisjComp$, which is computed in Line (13), includes the set of proper prime implicates of $c$ w.r.t. $\mathcal{P}(hist) \backslash \{\neg q \vee c\} \cup \{q\}$, which are in the target language, and in particular $m$. □

**Lemma 1** *Let $P$ be a set of clauses and $q$ a literal. Let $P' \subseteq P$ such that it contains clauses having a variable in common with $q$. If $m$ is a proper resolvent of $q$ w.r.t. $P$, then :*

- *either $m$ is a proper resolvent of $q$ w.r.t. $P'$,*
- *or the variable of $q$ is common with clauses of $P \backslash P'$,*
- *or there exists a clause $\neg q \vee c$ of $P'$ such that $c$ has variables in common with clauses of $P \backslash P'$ and $m$ is a proper resolvent of $c$ w.r.t. $P \backslash \{\neg q \vee c\} \cup \{q\}$.*

Proof: Let $m$ be a proper resolvent of $q$ w.r.t. $P$. If $m$ is different from $q$, there exists a clause $\neg q \vee c$ in $P$ such that $m$ is a proper resolvent of $c$ w.r.t. $P \cup \{q\}$. Since $P \backslash \{\neg q \vee c\} \cup \{q\} \equiv P \cup \{q\}$, $m$ is a proper resolvent of $c$ w.r.t. $P \backslash \{\neg q \vee c\} \cup \{q\}$.

- If such a clause does not exist in $P'$, it exists in $P \backslash P'$ and therefore the variable of $q$ is common with clauses of $P \backslash P'$.
- If there exists a clause $\neg q \vee c$ in $P'$ such that $m$ is a proper resolvent of $c$ w.r.t. $P \backslash \{\neg q \vee c\} \cup \{q\}$, and $m$ is not a proper resolvent of $q$ w.r.t. $P'$, then for every proof of $m$ there must exist a clause $c'$ in $P \backslash P'$ with which either $q$ or $\neg q \vee c$ must be resolved. Therefore, either $q$ or $c$ has variables in common with clauses of $P \backslash P'$. □

**Lemma 2** *Let $P$ be a set of clauses, and let $c = l_1 \vee \cdots \vee l_n$ be a clause. For every proper prime implicate $m$ of $c$ w.r.t. $P$, there exists $m_1, \ldots, m_n$ such that $m \equiv m_1 \vee \cdots \vee m_n$, and for every $i \in [1..n]$, $m_i$ is a proper prime implicate of $l_i$ w.r.t. $P$.*

Proof: Let $m$ be a proper prime implicate of $c$ w.r.t. $P$. For every literal $l_i$, let $Mod(l_i)$ be the set of models of $P$ which make $l_i$ true. If $Mod(l_i) = \emptyset$, that means that $\square$ is the only proper prime implicate of $l_i$ w.r.t. $P$. For every $i$ such that $Mod(l_i) \neq \emptyset$, every model in $Mod(l_i)$ is a model of $P \cup \{c\}$, and then a model of $m$ ; therefore, $m$ is a proper implicate of $l_i$ w.r.t. $P$, and, by definition of proper *prime* implicates, there exists a proper prime implicate $m_i$ of $l_i$ w.r.t. $P$ such that $m_i \models m$. Consequently, there exists $m_1, \ldots, m_n$ such that $m_1 \vee \cdots \vee m_n \models m$, and for every $i \in [1..n]$, $m_i$ is a proper prime implicate of $l_i$ w.r.t. $P$ ($m_i$ may be $\square$). Since $P \cup \{l_1 \vee \cdots \vee l_n\} \models m_1 \vee \cdots \vee m_n$ , and $m$ is a proper implicate of $l_1 \vee \cdots \vee l_n$ w.r.t. $P$, we necessarily get that $m \equiv m_1 \vee \cdots \vee m_n$. $\square$

### 3.2 Message-based Consequence Finding Algorithm

In this section, we exhibit the result of the transformation of the previous recursive algorithm into DeCA: a message-based DEcentralized Consequence finding Algorithm running locally on each peer. DeCA is composed of three procedures, each one being triggered by the reception of a message. The procedure ReceiveForthMessage is triggered by the reception of a $forth$ message $m(Sender, Receiver, forth, hist, l)$ sent by the peer $Sender$ to the peer $Receiver$ which executes the procedure: on the demand of $Sender$, with which it shares the variable of $l$, it processes the literal $l$. The procedure ReceiveBackMessage is triggered by the reception of a *back* message $m(Sender, Receiver, back, hist, r)$ sent by the peer $Sender$ to the peer $Receiver$ which executes the procedure: it processes the consequence $r$ (which is a clause the variables of which are target variables) sent back by $Sender$ for the literal $l$ (last added in the history) ; it may have to combine it with other consequences of literals being in the same clause as $l$. The procedure ReceiveFinalMessage is triggered by the reception of a $final$ message $m(Sender, Receiver, final, hist, true)$: the peer $Sender$ notifies the peer $Receiver$ that computation of the consequences of the literal $l$ (last added in the history) is completed. Those procedures handle two data structures stored at each peer: cons$(l, hist)$ caches the consequences of $l$ computed by the reasoning branch corresponding to $hist$ ; final$(q, hist)$ is set to true when the propagation of $q$ within the reasoning branch of the history $hist$ is completed.

The reasoning is initiated by the user (denoted by a particular peer $User$) sending to a given peer $P$ a message $m(User, P, forth, \emptyset, q)$. This triggers on the peer $P$ the local execution of the procedure ReceiveForthMessage$(m(User, P, forth, \emptyset, q))$. In the description of the procedures, since they are locally executed by the peer which receives the message, we will denote by $Self$ the receiver peer.

**Algorithm 2:** DeCA message passing procedure for propagating literals forth

ReceiveForthMessage($m(Sender, Self, forth, hist, q)$)
**if** $(\bar{q}, _, _) \in hist$
  **send** $m(Self, Sender, back, [(q, Self, \square)|hist], \square)$
  **send** $m(Self, Sender, final, [(q, Self, true)|hist], true)$
**else if** $q \in Self$ or $(q, Self, _) \in hist$
  **send** $m(Self, Sender, final, [(q, Self, true)|hist], true)$
**else**
  $\textsc{local}(Self) \leftarrow \{q\} \cup Resolvent(q, Self)$
  **if** $\square \in \textsc{local}(Self)$
    **send** $m(Self, Sender, back, [(q, Self, \square)|hist], \square)$
    **send** $m(Self, Sender, final, [(q, Self, true)|hist], true)$
  **else**
    $\textsc{local}(Self) \leftarrow \{c \in \textsc{local}(Self)|\ L(c) \in \mathcal{T}arget(Self)\}$
    **if** for every $c \in \textsc{local}(Self), S(c) = \square$
      **foreach** $c \in \textsc{local}(Self)$
        **send** $m(Self, Sender, back, [(q, Self, c)|hist], c)$
      **send** $m(Self, Sender, final, [(q, Self, true)|hist], true)$
    **else**
      **foreach** $c \in \textsc{local}(Self)$
        **if** $S(c) = \square$
          **send** $m(Self, Sender, back, [(q, Self, c)|hist], c)$
        **else**
          **foreach** literal $l \in S(c)$
            **if** $l \in \mathcal{T}arget(Self)$
              $\textsc{cons}(l, [(q, Self, c)|hist]) \leftarrow \{l\}$
            **else**
              $\textsc{cons}(l, [(q, Self, c)|hist]) \leftarrow \emptyset$
            $\textsc{final}(l, [(q, Self, c)|hist]) \leftarrow false$
            **foreach** $RP \in \textsc{acq}(l, Self)$
              **send** $m(Self, RP, forth, [(q, Self, c)|hist], l)$

**Algorithm 3:** DeCA message passing procedure for processing the return of consequences

ReceiveBackMessage($m(Sender, Self, back, hist, r)$)
$hist$ is of the form $[(l', Sender, c'), (q, Self, c)|hist']$
$\textsc{cons}(l', hist) \leftarrow \textsc{cons}\ (l', hist) \cup \{r\}$
$\textsc{result} \leftarrow \bigotimes_{l \in S(c) \backslash \{l'\}} \textsc{cons}(l, hist) \otimes \{L(c) \vee r\}$
**if** $hist' = \emptyset$, $U \leftarrow User$ **else** $U \leftarrow$ the first peer $P'$ of $hist'$
**foreach** $cs \in \textsc{result}$
  **send** $m(Self, U, back, [(q, Self, c)|hist'], cs)$

**Algorithm 4:** DeCA message passing procedure for notifying termination

ReceiveFinalMessage($m(Sender, Self, final, hist, true)$)
$hist$ is of the form $[(l', Sender, true), (q, Self, c)|hist']$
$\textsc{final}(l', hist) \leftarrow true$
**if** for every $l \in S(c)$, $\textsc{final}(l, hist) = true$
  **if** $hist' = \emptyset$ $U \leftarrow User$ **else** $U \leftarrow$ the first peer $P'$ of $hist'$
  **send** $m(Self, U, final, [(q, Self, true)|hist'], true)$
  **foreach** $l \in S(c)$
    $\textsc{cons}(l, [(l, Sender, _), (q, Self, c)|hist']) \leftarrow \emptyset$

The following theorem states two important results: first, the message-based distributed algorithm computes the same results as the algorithm of Section 3.1, and thus, is complete under the same conditions as in Theorem 2 ; second the user is notified of the termination when it occurs, which is crucial for an anytime algorithm.

**Theorem 3** *Let $r$ be a result returned by $RCF(q, P)$. If $P$ receives from the user the message $m(User, P, forth, \emptyset, q)$, then a message $m(P, User, back, [(q, P, _)], r)$ will be produced. If $r$ is the last result returned by $RCF(q, P)$, then the user will be notified of the termination by a message $m(P, User, final, [(q, P, true)], true)$.*

Proof: We prove by induction on the number of recursive calls of $RCFH(q, SP, hist)$ that:

(1) for any result $r$ returned by $RCFH(q, SP, hist)$, there exists $P \in SP$ such that $P$ is bound to send a message $m(P, S, back, [(q, P, _)|hist], r)$ after receiving the message $m(S, P, forth, hist, q)$,

(2) if $r$ is the last result returned by $RCFH(q, SP, hist)$, all the peers $P \in SP$ are bound to send the message $m(P, S, final, [(q, P, true)|hist], true)$, where $S$ is the first peer in the history.

• $rc = 0$: either one of the conditions of Lines (1), (2), (4) or (6) of the algorithm $RCFH(q, SP, hist)$ is satisfied. We have shown in the proof of Theorem 2 that if the conditions of Lines (2) and (4) are satisfied, $\square$ is the only result returned by the algorithm. The condition of Line (2) of the algorithm $RCFH(q, SP, hist)$ corresponds to the condition of Line (1) of the algorithm ReceiveForthMessage($m(S, P, forth, hist, q)$) for any $P$ of $SP$, which triggers the sending of a message $m(P, S, back, [(q, P, \square)|hist], \square)$ (Line (2)) and of a message $m(P, S, final, [(q, P, true)|hist], true)$ (Line(3)). If the condition of Line (4) of the algorithm $RCFH(q, SP, hist)$ is satisfied, there exists $P \in SP$ such that $\square \in P$. That condition corresponds to the condition of Line (8) of the algorithm for ReceiveForthMessage($m(S, P, forth, hist, q)$), which triggers the sending of a message $m(P, S, back, [(q, P, \square)|hist], \square)$ (Line (9) and of a message $m(P, S, final, [(q, P, true)|hist], true)$ (Line (10)). The condition (1) of the algorithm $RCFH(q, SP, hist)$, in which no result is returned (see proof of Theorem 2), corresponds to the condition of Line (4) of the algorithm ReceiveForthMessage($m(S, P, forth, hist, q)$), for every $P \in SP$, which only triggers the sending of a final message (Line (5)). Finally, if the condition of Line (6) of the algorithm $RCFH(q, SP, hist)$ is satisfied, there exists $P \in SP$ such that $r \in$ local($P$). The condition of Line (6) of the algorithm $RCFH(q, SP, hist)$ corresponds to the condition of Line (13) in ReceiveForthMessage($m(S, P, forth, hist, q)$), which triggers the sending of all the messages $m(P, S, back, [(q, P, c)|hist], c)$, where $c$ is a clause of local($P$) (Line (15)), and in particular the message $m(P, S, back, [(q, P, r)|hist], r)$. It triggers too the sending of a final message (Line (16)) for $P$. If $r$ is the last result returned by $RCFH(q, SP, hist)$, such final messages has been sent by every $P \in SP$.

• Suppose the induction hypothesis true for $rc \leq p$, and let $\Gamma = (\mathcal{P}, \text{acq})$ a P2PIS such that $RCFH(q, SP, hist)$ requires $p + 1$ recursive calls to terminate.

- If there exists $P \in SP$ such that $r \in$ local($P$), $r$ is not the last result returned by the algorithm, and $r$ is one of the clauses $c$ involved in the iteration of the loop of Line (18) of the algorithm ReceiveForthMessage($m(S, P, forth, hist, q)$), and verifying the condition of Line (19), which triggers the sending of the message $m(P, S, back, [(q, P, r)| hist], r)$ (Line (20)).

-If there exists $P \in SP$ and a clause $c : l_1 \vee \cdots \vee l_k \vee L(c)$ of $\textsc{local}(P)$ such that $c$ is involved in the iteration of the loop of Line (9) of the algorithm $RCFH(q, P, hist)$, and $r$ is an element $r_1 \vee \cdots \vee r_k \vee L(c)$ of $(\circledvee_{l \in S(c)} \textsc{answer}(l)) \circledvee \{L(c)\}$ computed at Line (12), where each $\textsc{answer}(l)$ is obtained as the result of $RCFH(l, \textsc{acq}(l, P), [(q, P, c)|hist])$ (Line (13)), which requires $p$ or less than $p$ recursive calls. By induction, for each literal $l_i \in S(c)$, there exists $RP_i \in \textsc{acq}(l_i, P)$ such that $RP_i$ sends a message $m(RP_i, P, back, [(l_i, RP_i, _), (q, P, c)|hist], r_i)$ if it has received the message $m(P, RP_i, forth, [(q, P, c)|hist], l_i)$. The loop of Line (11) of the algorithm $RCFH(q, SP, hist)$ corresponds to the loop of Line (22) of the algorithm ReceiveForthMessage($m(S, P, forth, hist, q)$), which triggers the sending of the messages $m(P, RP_i, forth, [(q, P, c)|hist], l_i)$ for each literal $l_i \in S(c)$ (Line (29)). Therefore, according to the induction hypothesis, for every $l_i \in S(c)$, $RP_i$ sends a message $m(RP_i, P, back, [(l_i, RP_i, _), (q, P, c)|hist], r_i)$. When the last of those messages (let us say $m(RP_j, P, back, [(l_j, RP_j, _), (q, P, c)|hist], r_j))$ is processed, $r$ is produced by Line (3) of ReceiveBackMessage($m(RP_j, P, back, [(l_j, RP_j, _), (q, P, c)|hist], r_j)$), and there exists a peer $U$ such that $P$ is bound to send to it the message $m(P, U, back, [(q, P, c)|hist], r)$ (Line (6)).

- If $r$ is the last result returned by the algorithm $RCFH(q, SP, hist)$, for every $P \in SP$, for every $c \in \textsc{local}(P)$, for every $l \in S(c)$, $RCFH(l, \textsc{acq}(l, P), [(q, P, c)|hist])$ has finished, and, by induction, every peer $RP$ of $\textsc{acq}(l, P)$ has sent a message $m(RP, P, final, [(l, RP, true), (q, P, c)|hist], true)$. Therefore, the condition of Line (3) of the algorithm ReceiveFinalMessage($m(RP, P, final, [(l, RP, _), (q, P, c)|hist], true)$) is satisfied, which triggers the sending of a message $m(P, U, final, [(q, P, true)|hist], true)$ (Line (5)).

□

For sake of simplicity, both recursive and distributed algorithms have been presented as applying to literals. It does not mean that the formulas that we consider are limited to literals. Clauses can be handled by splitting them into literals and then using the $\circledvee$ operator to recompose the results obtained for each literal.

It is also important to notice that □ can be returned by our algorithm as a proper prime implicate because of the lines (1) to (3) and (8) to (10) in ReceiveForthMessage. In that case, as a corollary of the above theorems, the P2PIS is detected unsatisfiable with the input clause. Therefore, our algorithm can be exploited for checking the satisfiability of the P2PIS at each join of a new peer theory.

## 4. Application to the Semantic Web: the somewhere Peer-to-peer Data Management System

The Semantic Web (Berners-Lee, Hendler, & Lassila, 2001) envisions a world wide distributed architecture where data and computational resources will easily inter-operate based on semantic marking up of web resources using *ontologies*. Ontologies are a formalization of the semantics of application domains (e.g., tourism, biology, medicine) through the definition of classes and relations modeling the domain objects and properties that are considered as meaningful for the application. Most of the concepts, tools and techniques deployed so far by the Semantic Web community correspond to the "big is beautiful" idea that high expressivity is needed for describing domain ontologies. As a result, when they are applied, the current Semantic Web technologies are mostly used for building thematic portals but do not scale up to the Web. In contrast, Somewhere promotes a "small is beautiful" vision of the Semantic Web (Rousset, 2004) based on simple personalized ontologies (e.g., taxonomies of atomic classes) but which are distributed at a large scale. In this vision of the Semantic Web introduced by Plu, Bellec, Agosto, and van de Velde (2003), no user imposes to others his own ontology but logical mappings between ontologies make possible the creation of a web of people in which personalized semantic marking up of data cohabits nicely with a collaborative exchange of data. In this view, the Web is a huge peer-to-peer data management system based on simple distributed ontologies and mappings.

Peer-to-peer data management systems have been proposed recently (Halevy et al., 2003b; Ooi, Shu, & Tan, 2003; Arenas, Kantere, Kementsietsidis, Kiringa, Miller, & Mylopoulos, 2003; Bernstein, Giunchiglia, Kementsietsidis, Mylopoulos, Serafini, & Zaihraheu, 2002; Calvanese et al., 2004) to generalize the centralized approach of information integration systems based on single mediators. In a peer-to-peer data management system, there is no central mediator: each peer has its own ontology and data or services, and can mediate with some other peers to ask and answer queries. The existing systems vary according to (a) the expressive power of their underlying data model and (b) the way the different peers are semantically connected. Both characteristics have impact on the allowed queries and their distributed processing.

In Edutella (Nejdl et al., 2002), each peer stores locally data (educational resources) that are described in RDF relative to some reference ontologies (e.g., Dmoz - http://dmoz.org). For instance, a peer can declare that it has data related to the concept of the dmoz taxonomy corresponding to the path *Computers/Programming/Languages/Java*, and that for such data it can export the *author* and the *date* properties. The overlay network underlying Edutella is a hypercube of super-peers to which peers are directly connected. Each super-peer is a mediator over the data of the peers connected to it. When it is queried, its first task is to check if the query matches with its schema: if that is the case, it transmits the query to the peers connected to it, which are likely to store the data answering the query ; otherwise, it routes the query to some of its neighbour super-peers according to a strategy exploiting the hypercube topology for guaranteeing a worst-case logarithmic time for reaching the relevant super-peer.

In contrast with Edutella, Piazza (Halevy et al., 2003b, 2003a) does not consider that the data distributed over the different peers must be described relatively to some existing reference schemas. In Piazza, each peer has its own data and schema and can mediate with some other peers by declaring *mappings* between its schema and the schemas of those peers. The topology of the network is not fixed (as the hypercube in Edutella) but accounts for the existence of mappings between peers: two peers are logically connected if there exists a mapping between their two schemas. The underlying data model of the first version of Piazza (Halevy et al., 2003b) is relational and the mappings between relational peer schemas are inclusion or equivalence statements between conjunctive queries. Such a mapping formalism encompasses the *local-as-views* and the *global-as-views* (Halevy, 2000) formalisms used in information integration systems based on single mediators. The price to pay is that query answering is undecidable except if some restrictions are imposed on the mappings or on the topology of the network (Halevy et al., 2003b). The currently implemented version of Piazza (Halevy et al., 2003a) relies on a tree-based data model: the data is in XML and the mappings are equivalence and inclusion statements between XML queries. Query answering is implemented based on practical (but not complete) algorithms for XML query containment and rewriting. The scalability of Piazza so far does not go up to more than about 80 peers in the published experiments and relies on a wide range of optimizations (mappings composition, Madhavan & Halevy, 2003, paths pruning, Tatarinov & Halevy, 2004), made possible by the centralized storage of all the schemas and mappings in a global server.

In Somewhere, we have made the choice of being fully distributed: there are neither super-peers (as in Edutella) nor a central server having the global view of the overlay network (as in Piazza). In addition, we aim at scaling up to thousands of peers. To make it possible, we have chosen a simple class-based data model in which the data is a set of resource identifiers (e.g., URIs), the schemas are (simple) definitions of classes possibly constrained by inclusion, disjunction or equivalence statements, and mappings are inclusion, disjunction or equivalence statements between classes of different peer schemas. That data model is in accordance with the W3C recommendations since it is captured by the propositional fragment of the OWL ontology language (http://www.w3.org/TR/owl-semantics). Note that OWL makes possible, through a declarative import mechanism, to retrieve ontologies that are physically distributed. Using this transitive mechanism in peer data management systems amounts in the worst case to centralized on a single peer the whole set of peer ontologies, and to reason locally. Our feeling is that on very large networks such a mechanism cannot scale up satisfactorily. Moreover, because of the dynamicity of peer-to-peer settings, such imports would have to be re-actualized each time that a peer joins or quit the network.

Section 4.1 defines the Somewhere data model, for which an illustrative example is given in Section 4.2. In Section 4.3, we show how query rewriting in Somewhere, and thus query answering, can be reduced by a propositional encoding to distributed reasoning in propositional logic.

## 4.1 Somewhere Data model

In Somewhere, a new peer joins the network through some peers that it knows (its acquaintances) by declaring mappings between its own ontology and the ontologies of its acquaintances. Queries are posed to a given peer using its local ontology. The answers that are expected are not only instances of local classes but possibly instances of classes of peers distant from the queried peer if it can be infered from the peer ontologies and the mappings that they satisfy the query. Local ontologies, storage descriptions and mappings are defined using a fragment of OWL DL which is the description logic fragment of the Ontology Web Language recommended by W3C. We call OWL PL the fragment of OWL DL that we consider in Somewhere, where PL stands for propositional logic. OWL PL is the fragment of OWL DL reduced to the disjunction, conjunction and negation constructors for building class descriptions.

### 4.1.1 Peer ontologies

Each peer ontology is made of a set of class definitions and possibly a set of equivalence, inclusion or disjointness axioms between class descriptions. A class description is either the universal class ($\top$), the empty class ($\bot$), an atomic class or the union ($\sqcup$), intersection ($\sqcap$) or complement ($\neg$) of class descriptions.

The name of atomic classes is unique to each peer: we use the notation $P{:}A$ for identifying an atomic class $A$ of the ontology of a peer $P$. The *vocabulary* of a peer $P$ is the set of names of its atomic classes.

**Class descriptions**

| | *Logical notation* | *OWL notation* |
|---|---|---|
| universal class | $\top$ | $Thing$ |
| empty class | $\bot$ | $Nothing$ |
| atomic class | $P{:}A$ | $classID$ |
| conjunction | $D1 \sqcap D2$ | $intersectionOf(D1\ D2)$ |
| disjunction | $D1 \sqcup D2$ | $unionOf(D1\ D2)$ |
| negation | $\neg D$ | $complementOf(D)$ |

**Axioms of class definitions**

| | *Logical notation* | *OWL notation* |
|---|---|---|
| Complete | $P{:}A \equiv D$ | $Class(P{:}A\ complete\ D)$ |
| Partial | $P{:}A \sqsubseteq D$ | $Class(P{:}A\ partial\ D)$ |

**Axioms on class descriptions**

| | *Logical notation* | *OWL notation* |
|---|---|---|
| equivalence | $D1 \equiv D2$ | $EquivalentClasses(D1\ D2)$ |
| inclusion | $D1 \sqsubseteq D2$ | $SubClassOf(D1\ D2)$ |
| disjointness | $D1 \sqcap D2 \equiv \bot$ | $DisjointClasses(D1\ D2)$ |

Taxonomies of atomic classes (possibly enriched by disjointness statements between atomic classes) are particular cases of the allowed ontologies in Somewhere . Their specification is made of a set of inclusion (and disjointness) axioms involving atomic classes only: there is no class definition using (conjunction, disjunction or negation) constructors.

#### 4.1.2 Peer storage descriptions

The specification of the data that is stored locally in a peer $P$ is done through the declaration of atomic *extensional classes* defined in terms of atomic classes of the peer ontology, and assertional statements relating data identifiers (e.g., URIs) to those extensional classes. We restrict the axioms defining the extensional classes to be inclusion statements between an atomic extensional class and a description combining atomic classes of the ontology. We impose that restriction in order to fit with a *Local-as-Views* approach and an open-world assumption within the information integration setting (Halevy, 2000). We will use the notation $P{:}ViewA$ to denote an extensional class $ViewA$ of the peer $P$.

**Storage description**

declaration of extensional classes:

| *Logical notation* | *OWL notation* |
|---|---|
| $P{:}ViewA \sqsubseteq C$ | $SubClassOf(P{:}ViewA\ \ C)$ |

assertional statements:

| *Logical notation* | *OWL notation* |
|---|---|
| $P{:}ViewA(a)$ | $individual(a\ type(P{:}ViewA))$ |

#### 4.1.3 Mappings

Mappings are disjointness, equivalence or inclusion statements involving atomic classes of different peers. They express the semantic correspondence that may exist between the ontologies of different peers.

#### 4.1.4 Schema of a Somewhere network

In a Somewhere network, the schema is not centralized but distributed through the union of the different peer ontologies and the mappings. The important point is that each peer has a partial knowledge of the schema: it just knows its own local ontology and the mappings with its acquaintances.

Let $\mathcal{P}$ be a Somewhere peer-to-peer network made of a collection of peers $\{P_i\}_{i=1..n}$. For each peer $P_i$, let $O_i$, $V_i$ and $M_i$ be the sets of axioms defining respectively the local ontology of $P_i$, the declaration of its extensional classes and the set of mappings stated at $P_i$ between classes of $O_i$ and classes of the ontologies of the acquaintances of $P_i$. The schema of $\mathcal{P}$, denoted $S(\mathcal{P})$, is the union $\bigcup_{i=1..n} O_i \cup V_i \cup M_i$ of the ontologies, the declaration of extensional classes and of the sets of mappings of all the peers.

#### 4.1.5 Semantics

The semantics isis the standard semantics of first order logic defined in terms of *interpretations*. An interpretation $I$ is a pair $(\Delta^I, .^I)$ where $\Delta$ is a non-empty set, called the domain of interpretation, and $.^I$ is an interpretation function which assigns a subset of $\Delta^I$ to every class identifier and an element of $\Delta^I$ to every data identifier.

An interpretation $I$ is a *model* of the distributed schema of a Somewhere peer-to-peer network $\mathcal{P} = \{P_i\}_{i=1..n}$ iff for each axiom in $\bigcup_{i=1..n} O_i \cup V_i \cup M_i$ is satisfied by $I$. Interpretations of axioms rely on interpretations of class descriptions which are inductively defined as follows:

- $\top^I = \Delta^I$, $\bot^I = \emptyset$
- $(C_1 \sqcup C_2)^I = C_1^I \cup C_2^I$
- $(C_1 \sqcap C_2)^I = C_1^I \cap C_2^I$
- $(\neg C)^I = \Delta^I \backslash C^I$

Axioms are satisfied if the following holds:

- $C \sqsubseteq D$ is satisfied in $I$ iff $C^I \subseteq D^I$
- $C \equiv D$ is satisfied in $I$ iff $C^I = D^I$
- $C \sqcap D \equiv \bot$ is satisfied in $I$ iff $C^I \cap D^I = \emptyset$

A Somewhere peer-to-peer network is *satisfiable* iff its (distributed) schema has a model.

Given a Somewhere peer-to-peer network $\mathcal{P} = \{P_i\}_{i=1..n}$, a class description $C$ *subsumes* a class description $D$ iff for any model $I$ of $S(\mathcal{P})$ $D^I \subseteq C^I$.

### 4.2 Illustrative Example

We illustrate the Somewhere data model on a small example of four peers modeling four persons Ann, Bob, Chris and Dora, each of them bookmarking URLs about restaurants they know or like, according to their own taxonomy for categorizing restaurants.

**Ann**, who is working as a restaurant critic, organizes its restaurant URLs according to the following classes:

• the class $Ann{:}G$ of restaurants considered as offering a "good" cooking, among which she distinguishes the subclass $Ann{:}R$ of those which are rated: $Ann{:}R \sqsubseteq Ann{:}G$

• the class $Ann{:}R$ is the union of three disjoint classes $Ann{:}S1$, $Ann{:}S2$, $Ann{:}S3$ corresponding respectively to the restaurants rated with $1, 2$ or $3$ stars:

$Ann{:}R \equiv Ann{:}S1 \sqcup Ann{:}S2 \sqcup Ann{:}S3$

$Ann{:}S1 \sqcap Ann{:}S2 \equiv \bot \quad Ann{:}S1 \sqcap Ann{:}S3 \equiv \bot$

$Ann{:}S2 \sqcap Ann{:}S3 \equiv \bot$

• the classes $Ann{:}I$ and $Ann{:}O$, respectively corresponding to Indian and Oriental restaurants

• the classes $Ann{:}C$, $Ann{:}T$ and $Ann{:}V$ which are subclasses of $Ann{:}O$ denoting Chinese, Taï and Vietnamese restaurants respectively: $Ann{:}C \sqsubseteq Ann{:}O$, $Ann{:}T \sqsubseteq Ann{:}O$, $Ann{:}V \sqsubseteq Ann{:}O$

Suppose that the data stored by Ann that she accepts to make available deals with restaurants of various specialties, and only with those rated with 2 stars among the rated restaurants. The extensional classes declared by Ann are then:
$Ann{:}ViewS2 \sqsubseteq Ann{:}S2$, $Ann{:}ViewC \sqsubseteq Ann{:}C$, $Ann{:}ViewV \sqsubseteq Ann{:}V$,
$Ann{:}ViewT \sqsubseteq Ann{:}T$, $Ann{:}ViewI \sqsubseteq Ann{:}I$

**Bob**, who is fond of Asian cooking and likes high quality, organizes his restaurant URLs according to the following classes:

• the class $Bob{:}A$ of Asian restaurants

• the class $Bob{:}Q$ of high quality restaurants that he knows

Suppose that he wants to make available every data that he has stored. The extensional classes that he declares are $Bob{:}ViewA$ and $Bob{:}ViewQ$ (as subclasses of $Bob{:}A$ and $Bob{:}Q$):
$Bob{:}ViewA \sqsubseteq Bob{:}A$, $Bob{:}ViewQ \sqsubseteq Bob{:}Q$

**Chris** is more fond of fish restaurants but recently discovered some places serving a very nice cantonese cuisine. He organizes its data with respect to the following classes:

• the class $Chris{:}F$ of fish restaurants,

• the class $Chris{:}CA$ of Cantonese restaurants

Suppose that he declares the extensional classes $Chris{:}ViewF$ and $Chris{:}ViewCA$ as subclasses of $Chris{:}F$ and $Chris{:}CA$ respectively:
$Chris{:}ViewF \sqsubseteq Chris{:}F$, $Chris{:}ViewCA \sqsubseteq Chris{:}CA$

**Dora** organizes her restaurants URLs around the class $Dora{:}DP$ of her preferred restaurants, among which she distinguishes the subclass $Dora{:}P$ of pizzerias and the subclass $Dora{:}SF$ of seafood restaurants.
Suppose that the only URLs that she stores concerns pizzerias: the only extensional class that she has to declare is $Dora{:}ViewP$ as a subclass of $Dora{:}P$: $Dora{:}ViewP \sqsubseteq Dora{:}P$

**Ann**, **Bob**, **Chris** and **Dora** express what they know about each other using mappings stating properties of class inclusion or equivalence.

**Ann** is very confident in Bob's taste and agrees to include Bob' selection as good restaurants by stating $Bob{:}Q \sqsubseteq Ann{:}G$. Finally, she thinks that Bob's Asian restaurants encompass her Oriental restaurant concept: $Ann{:}O \sqsubseteq Bob{:}A$

**Bob** knows that what he calls Asian cooking corresponds exactly to what Ann classifies as Oriental cooking. This may be expressed using the equivalence statement : $Bob{:}A \equiv Ann{:}O$ (note the difference of perception of Bob and Ann regarding the mappings between $Bob{:}A$ and $Ann{:}O$)

**Chris** considers that what he calls fish specialties is a particular case of Dora seafood specialties: $Chris{:}F \sqsubseteq Dora{:}SF$

**Dora** counts on both Ann and Bob to obtain good Asian restaurants : $Bob{:}A \sqcap Ann{:}G \sqsubseteq Dora{:}DP$

Figure 2 describes the resulting overlay network. In order to alleviate the notations, we omit the local peer name prefix except for the mappings. Edges are labeled with the class identifiers that are shared through the mappings between peers.

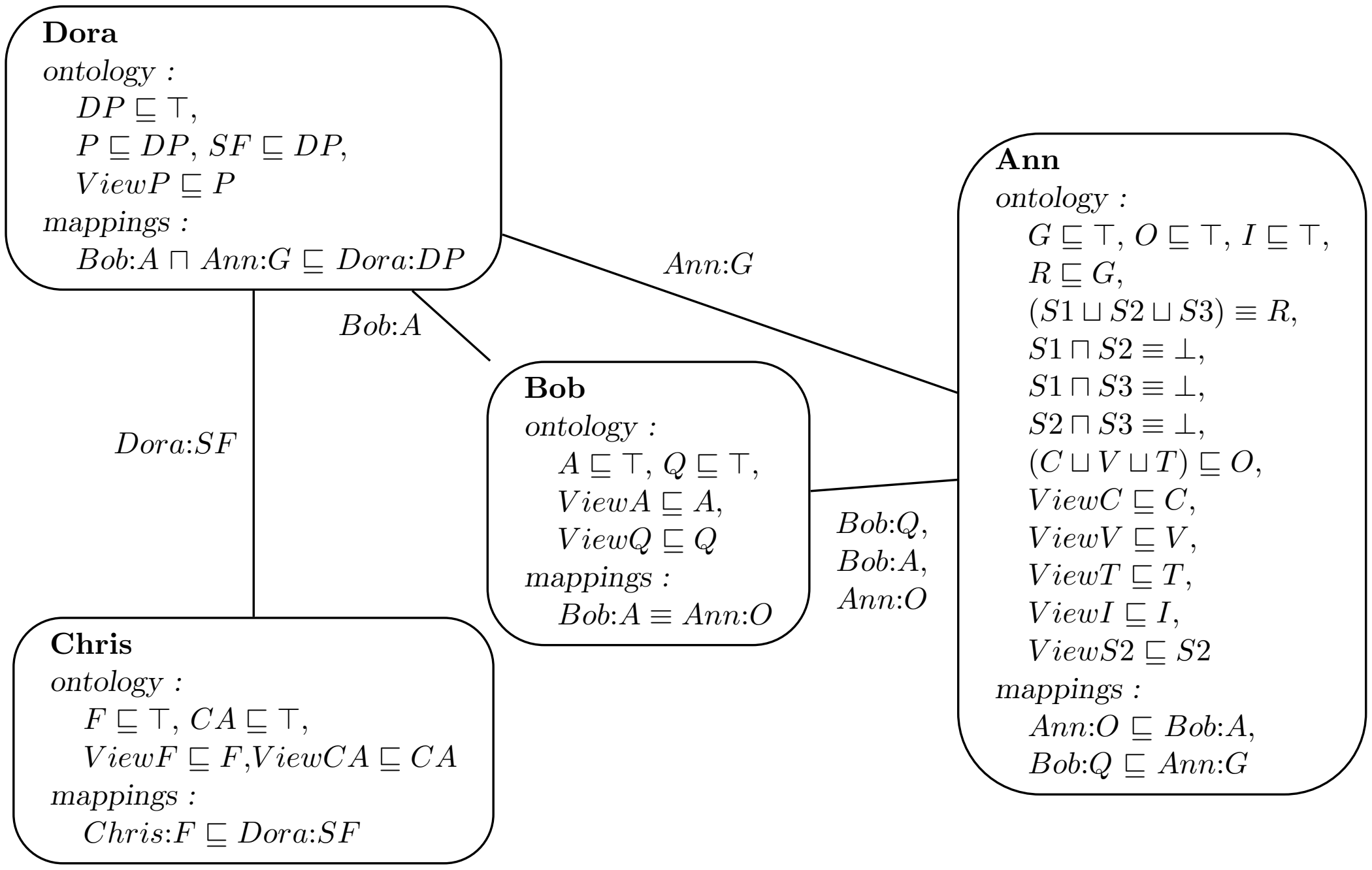

Figure 2: The restaurants network

### 4.3 Query Rewriting in Somewhere through Propositional Encoding

In Somewhere, each user interrogates the peer-to-peer network through one peer of his choice, and uses the vocabulary of this peer to express his query. Therefore, queries are logical combinations of classes of a given peer ontology.

The corresponding answer sets are expressed in intention in terms of the combinations of extensional classes that are *rewritings* of the query. The point is that extensional classes of several distant peers can participate to the rewritings, and thus to the answer of a query posed to a given peer.

**Definition 6 (Rewritings)** *Given a* SOMEWHERE *peer-to-peer network* $\mathcal{P} = \{P_i\}_{i=1..n}$*, a logical combination* $Q_e$ *of extensional classes is a* rewriting *of a query* $Q$ *iff* $Q$ *subsumes* $Q_e$ *w.r.t.* $\mathcal{P}$*.*
$Q_e$ *is a* proper rewriting *if there exists some model of* $I$ *of* $S(\mathcal{P})$ *such that* $Q_e^I \neq \emptyset$*.* $Q_e$ *is a* conjunctive rewriting *if it is a rewriting which is a conjunction of extensional classes.*
$Q_e$ *is a* maximal (conjunctive) rewriting *if there does not exist another (conjunctive) rewriting* $Q'_e$ *of* $Q$ *(strictly) subsuming* $Q_e$ *w.r.t.* $\mathcal{P}$*.*

In general, finding all answers in a peer data management system is a critical issue (Halevy et al., 2003b). In our setting however, we are in a case where all the answers can be obtained using *rewritings* of the query: it has been shown (Goasdoué & Rousset, 2004) that when a query has a *finite number of maximal conjunctive rewritings*, then *all* its answers (a.k.a. certain answers) can be obtained as the union of the answer sets of its rewritings. From the query answering point of view, it is the notion of *proper* rewriting which is relevant because it guarantees a non empty set of answers. If a query has no proper rewriting, it won't get any answer.

In the SOMEWHERE setting, query rewriting can be equivalently reduced to distributed reasoning over logical propositional theories by a straighforward propositional encoding of the query and of the distributed schema of a SOMEWHERE network. It consists in transforming each query and schema statement into a propositional formula using class identifiers as propositional variables.

The propositional encoding of a class description $D$, and thus of a query, is the propositional formula $Prop(D)$ obtained inductively as follows:

- $Prop(\top) = true$, $Prop(\bot) = false$
- $Prop(A) = A$, if $A$ is an atomic class
- $Prop(D_1 \sqcap D_2) = Prop(D_1) \wedge Prop(D_2)$
- $Prop(D_1 \sqcup D_2) = Prop(D_1) \vee Prop(D_2)$
- $Prop(\neg D) = \neg(Prop(D))$

The propositional encoding of the schema $\mathcal{S}$ of a SOMEWHERE peer-to-peer network $\mathcal{P}$ is the distributed propositional theory $Prop(\mathcal{S})$ made of the formulas obtained inductively from the axioms in $\mathcal{S}$ as follows:

- $Prop(C \sqsubseteq D) = Prop(C) \Rightarrow Prop(D)$
- $Prop(C \equiv D) = Prop(C) \Leftrightarrow Prop(D)$
- $Prop(C \sqcap D \equiv \bot) = \neg Prop(C) \vee \neg Prop(D)$

From now on, for simplicity, we use the propositional clausal form notation for the queries and SOMEWHERE peer-to-peer network schemas. As an illustration, let us consider the propositional encoding of the example presented in Section 4.2. After application of the transformation rules, conversion of each produced formula in clausal form and suppression of tautologies, we obtain (Figure 3) a new acquaintance graph where each peer schema is described as a propositional theory.

We now state two propositions showing that query rewriting in SOMEWHERE can be reduced to consequence finding in a P2PIS as presented in the previous sections.

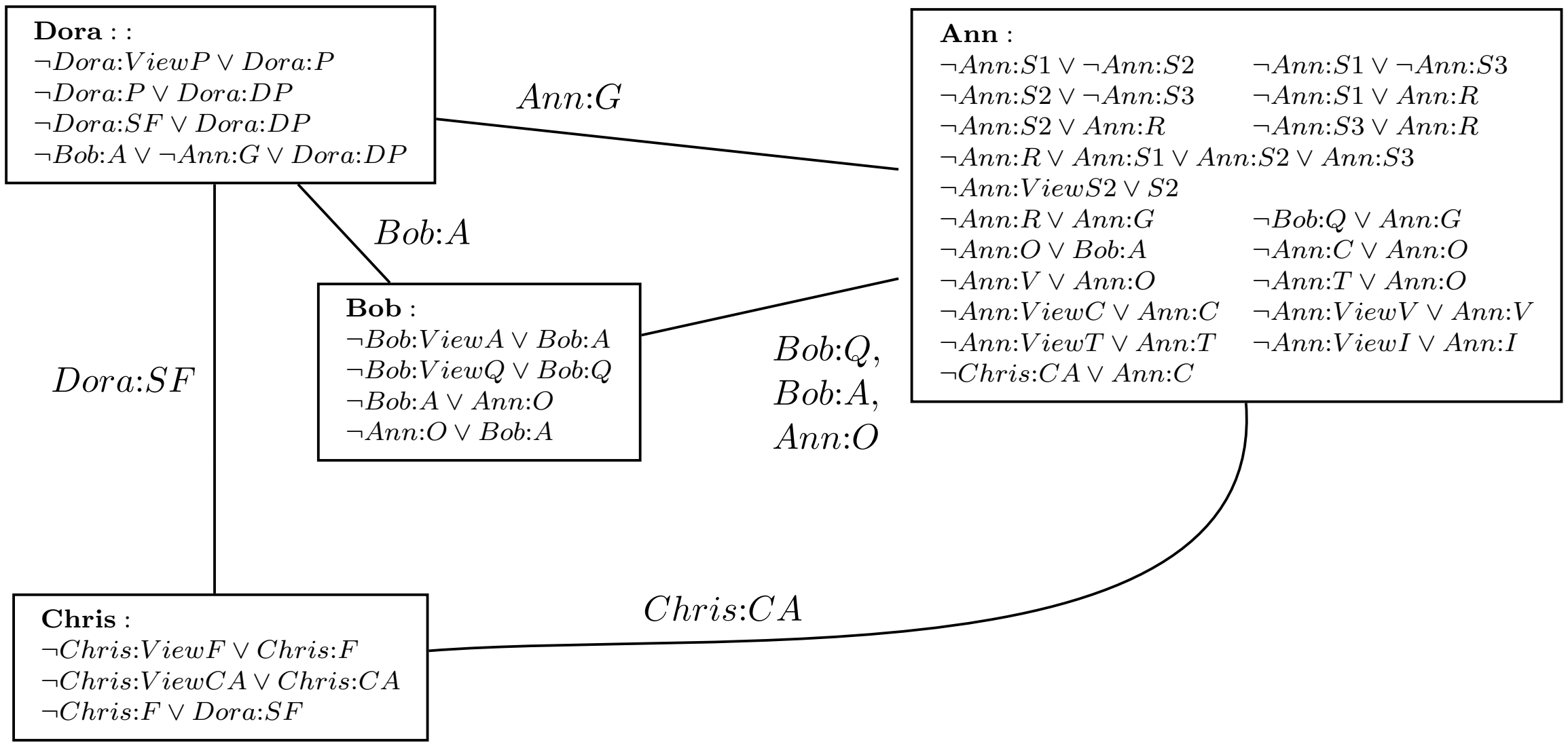

Figure 3: The propositional encoding for the restaurant network

Proposition 1 states that the propositional encoding transfers satisfiability and establishes the connection between proper (maximal) conjunctive rewritings and proper (prime) implicates.

Proposition 2 states that the P2PIS resulting from the propositional encoding of a Somewhere schema fulfills by construction the property exhibited in Theorem 2 as a sufficient condition for the completeness of the algorithms described in Section 3 that compute proper prime implicates of a clause w.r.t distributed propositional theories.

**Proposition 1** *Let $\mathcal{P}$ be a* Somewhere *peer-to-peer network and let $Prop(S(\mathcal{P}))$ be the propositional encoding of its schema. Let $V_e$ be the set of all the extensional classes.*

- *$S(\mathcal{P})$ is satisfiable iff $Prop(S(\mathcal{P}))$ is satisfiable.*
- *$q_e$ is a proper maximal conjunctive rewriting of a query $q$ iff $\neg Prop(q_e)$ is a* proper prime implicate of $\neg Prop(q)$ w.r.t. $Prop(S(\mathcal{P}))$ *such that all its variables are extensional classes.*

Proof: We first exhibit some properties that will be used in the proof of the proposition. Let $\mathcal{P}$ be a Somewhere network, $S(\mathcal{P})$ its schema and $Prop(S(\mathcal{P}))$ its propositional encoding. For an interpretation $I$ of $S(\mathcal{P})$, and an element $o$ of its domain of interpretation $\Delta^I$, we define the interpretation $p_o(I)$ of $Prop(S(\mathcal{P}))$ as follows: for every propositional variable $v$ of $Prop(S(\mathcal{P}))$ ($v$ is the name of an atomic class of $S(\mathcal{P})$), $v^{p_o(I)} = true$ iff $o \in v^I$. For an interpretation $J$ of $Prop(S(\mathcal{P}))$, we define the interpretation $i(J)$ of $S(\mathcal{P})$ as follows: the domain of $i(J)$ is $\{true\}$, and for every atomic class $A$ of $S(\mathcal{P})$ ($A$ is the name of a propositional variable of $Prop(S(\mathcal{P}))$), if $A^J = true$ then $A^{i(J)} = \{true\}$ else $A^{i(J)} = \emptyset$. It is easy to show the following properties, for every interpretation $I$ of $S(\mathcal{P})$ (and every $o \in \Delta^I$), for every interpretation $J$ of $Prop(S(\mathcal{P}))$:

1. for every class description $C$ of $S(\mathcal{P})$:

*a)* $o \in C^I \Leftrightarrow Prop(C)^{p_o(I)} = true$

*b)* $true \in C^{i(J)} \Leftrightarrow Prop(C)^J = true$

2. $I$ is a model of $S(\mathcal{P}) \Leftrightarrow p_o(I)$ is a model of $Prop(S(\mathcal{P}))$

3. $i(J)$ is a model of $S(\mathcal{P}) \Leftrightarrow J$ is a model of $Prop(S(\mathcal{P}))$.

- Suppose that $S(\mathcal{P})$ is satisfiable and let $I$ be a model of it and let $o$ be an element of $\Delta^I$: according to the above Property 2, $p_o(I)$ is a model of $Prop(S(\mathcal{P}))$, and thus $Prop(S(\mathcal{P}))$ is satisfiable. In the converse way, let $J$ be a propositional model of $Prop(S(\mathcal{P}))$. According to the above Property 3, $i(J)$ is a model of $S(\mathcal{P})$, and thus $S(\mathcal{P})$ is satisfiable.

- Suppose that $Prop(\neg q_e)$ is a proper prime implicate of $Prop(\neg q)$ w.r.t. $Prop(S(\mathcal{P}))$, such that all its variables are extensional classes, and let us show that $q_e$ is a proper maximal conjunctive rewriting of $q$.

  Let us first show that $q_e$ is a *rewriting* of $q$. Suppose that it is not the case: there exists an interpretation $I$ of $S(\mathcal{P})$ such that $q_e^I \not\subseteq q^I$ and thus an element of $o \in \Delta^I$ such that $o \in q_e^I$ and $o \notin q^I$. According to Property 2, $p_o(I)$ is a model of $Prop(S(\mathcal{P}))$, and according to Property 1.a, $(Prop(q_e))^{p_o(I)} = true$ and $(Prop(q))^{p_o(I)} \neq true$, i.e., $(Prop(\neg q_e))^{p_o(I)} = false$ and $(Prop(\neg q))^{p_o(I)} = true$. This is impossible since it would contradict the fact that $Prop(\neg q_e)$ is a proper prime implicate of $\{Prop(\neg q)\} \cup Prop(S(\mathcal{P}))$. Therefore, $q_e$ must be a rewriting of $q$.

  Let us show now that $q_e$ is a *conjunctive* rewriting of $q$, i.e., $q_e$ is a conjunction of extensional classes. In $S(\mathcal{P})$, extensional classes only appear in inclusion axioms. Therefore, the propositional encoding of $S(\mathcal{P})$ ensures that extensional classes only appear as variables in $Prop(S(\mathcal{P}))$ in negative literals. Resolution being sound and complete for prime implicate computation, $Prop(\neg q_e)$ is obtained as the result of a finite chain of resolutions, starting from the clauses of $Prop(S(\mathcal{P}))$ and from $Prop(\neg q)$. Since in $Prop(S(\mathcal{P}))$ extensional classes only appear in negative literals, they only appear in negative literals in the computed implicates, and in particular in $Prop(\neg q_e)$. Therefore, $q_e$ is a conjunction of extensional classes.

  Let us show now that $q_e$ is a *proper* rewriting of $q$, i.e., that it is satisfiable in some model of $S(\mathcal{P})$. Since $Prop(\neg q_e)$ is a *proper* prime implicate of $Prop(\neg q)$ w.r.t. $Prop(S(\mathcal{P}))$, there exists a model $J$ of $Prop(S(\mathcal{P}))$ s.t. $Prop(\neg q_e)^J = false$ and thus $Prop(q_e)^J = true$. By Property 3, $i(J)$ is a model of $S(\mathcal{P})$ and according to the above Property 1.b, $true \in (q_e)^{i(J)}$. Therefore there exists a model of $S(P)$ in which $q_e$ is satisfiable, and thus $q_e$ is a proper rewriting of $q$.

  Finally, let us show that $q_e$ is a *maximal* conjunctive rewriting of $q$. Suppose that this is not the case. Then, there exists a conjunctive rewriting $q'_e$ of $q$ such that $q_e \models q'_e$ and $q_e \not\equiv q'_e$. This means that there exists an interpretation $I$ and an element $o \in \Delta^I$ such that $o \in {q'}_e^I$, thus $o \in q^I$, and $o \notin q_e^I$. According to Property 1.a, $Prop(q'_e)^{p_o(I)} = true$, $Prop(q)^{p_o(I)} = true$, and $Prop(q_e)^{p_o(I)} = false$, i.e., $Prop(\neg q'_e)^{p_o(I)} = false$, $Prop(\neg q)^{p_o(I)} = false$, and $Prop(\neg q_e)^{p_o(I)} = true$. This is impossible since it contradicts that $Prop(\neg q_e)$ is a prime implicate of $Prop(\neg q)$ w.r.t. $Prop(S(\mathcal{P}))$. Therefore, $q_e$ is a maximal conjunctive rewriting of $q$.

- Let us now prove the converse direction. Suppose that $q_e$ is a proper maximal conjunctive rewriting of a query $q$ and let us show hat $Prop(\neg q_e)$ is a proper prime implicate of $Prop(\neg q)$ w.r.t. $Prop(S(\mathcal{P}))$. By definition of a proper rewriting, for every model $I$ of $S(\mathcal{P})$ $q_e^I \subseteq q^I$, or equivalently $(\neg q)^I \subseteq (\neg q_e)^I$ and there exists a model $I'$ of $S(\mathcal{P})$ such that $q_e^{I'} \neq \emptyset$.

  Let us first show that $Prop(\neg q_e)$ is an *implicate* of $Prop(\neg q)$ w.r.t $Prop(S(\mathcal{P}))$. Suppose that this is not the case, i.e., that $\{Prop(\neg q)\} \cup Prop(S(\mathcal{P})) \not\models Prop(\neg q_e)$. Then, there exists a model $J$ of $\{Prop(\neg q)\} \cup Prop(S(\mathcal{P}))$ such that $Prop(\neg q_e)^J = false$. According to the above Property 3, $i(J)$ is a model of $S(\mathcal{P})$. According to the above Property 1.b, $true \in (\neg q)^{i(J)}$ and $(\neg q_e)^{i(J)} = \emptyset$. This is impossible since it contradicts that $(\neg q)^{i(J)} \subseteq (\neg q_e)^{i(J)}$. Therefore, $Prop(\neg q_e)$ is an implicate of $Prop(\neg q)$ w.r.t. $Prop(S(\mathcal{P}))$.

  Let us now show that $Prop(\neg q_e)$ is a *proper* implicate of $Prop(\neg q)$ w.r.t. $Prop(S(\mathcal{P}))$, i.e., that $Prop(\neg q_e)$ is not an implicate of $Prop(S(\mathcal{P}))$ alone. By definition of a proper rewriting, there exists a model $I'$ of $S(\mathcal{P})$ such that $q_e^{I'} \neq \emptyset$. Let $o$ be an element of $q_e^{I'}$. According to Property 2, $p_o(I')$ is a model of $Prop(S(\mathcal{P}))$, and according to Property 1.a, $(Prop(q_e))^{p_o(I')} = true$, i.e., $(Prop(\neg q_e))^{p_o(I')} = false$. Therefore, $Prop(\neg q_e)$ is not an implicate of $Prop(S(\mathcal{P}))$.

  Finally, let us show that $Prop(\neg q_e)$ is a *prime* implicate of $Prop(\neg q)$ w.r.t. $Prop(S(\mathcal{P}))$.

  Let us show that if $c$ is a clause such that $Prop(S(P)) \cup \{Prop(\neg q)\} \models c$ and $c \models Prop(\neg q_e)$, then $c \equiv Prop(\neg q_e)$. Since $c \models Prop(\neg q_e)$ and $Prop(\neg q_e)$ is a disjunction of negation of extensional classes, $c$ is a disjunction of a subset of the literals of $Prop(\neg q_e)$. Let $q'_e$ be the conjunction of the extensional classes of $c$, then $c = Prop(\neg q'_e)$. We have proved previously that if $Prop(\neg q'_e)$ is an implicate of $Prop(\neg q)$ w.r.t. $Prop(S(P))$ then $q'_e$ is a rewriting of $q$, and similarly that if $Prop(\neg q_e)$ is an implicate of $Prop(\neg q'_e)$ w.r.t. $Prop(S(P))$ then $q'_e$ subsumes $q_e$. Therefore, $q'_e$ is a rewriting of $q_e$ which subsumes $q_e$. Since $q_e$ is a maximal conjunctive rewriting of q, $q'_e \equiv q_e$, thus $\neg q'_e \equiv \neg q_e$ and $Prop(\neg q'_e) \equiv Prop(\neg q_e)$, i.e. $c \equiv Prop(\neg q_e)$.

□

**Proposition 2** *Let $\mathcal{P}$ be a* Somewhere *peer-to-peer network and let $Prop(S(\mathcal{P}))$ be the propositional encoding of its schema. Let $\Gamma = (Prop(S(\mathcal{P})), \text{ACQ})$ be the corresponding P2PIS, where the set of labelled edges* ACQ *is such that: (P":A, P, P') $\in$* ACQ *iff P":A is involved in a mapping between P and P' (i.e, P" = P or P" = P'). For every $P$, $P'$ and $v \in \mathcal{V}_P \cap \mathcal{V}_{P'}$ there exists a path between $P$ and $P'$ in $\Gamma$ such that all edges of it are labelled with $v$.*

Proof: Let $P$ and $P'$ be two peers having a variable $v$ in common. Since the vocabularies of the local ontologies of different peers are disjoint, $v$ is necessarily a variable $P''{:}A$ involved in a mapping declared between $P$ and some its acquaintances $P_1$ (and thus $P'' = P$ or $P''= P_1$), or between $P'$ and some of its acquaintances $P'_1$ (in this case $P'' = P'$ or $P''= P'_1$).
- If $v$ is of the form $P''{:}A$ such that $P'' = P$ (respectively $P'' = P'$), then $P{:}A$ is an atomic class of $P$ (respectively $P'{:}A$ is an atomic class of $P'$) which is involved in a mapping between $P$ and $P'$, and therefore, there is an edge (and thus a path) between $P$ and $P'$ labelled with $v$ ($P{:}A$ or $P'{:}A$ respectively) in $\Gamma$.
- If $v$ is of the form $P''{:}A$ such that $P''$ is distinct from $P$ and $P'$, then $P''{:}A$ is an atomic class of $P''$, which is involved in a mapping between $P''$ and $P$ and in a mapping between $P''$ and $P'$. Therefore, there exists an edge between $P''$ and $P$ labelled with $v$ and an edge between $P''$ and $P'$ labelled with $v$, and thus a path between $P$ and $P'$ such that all edges of it are labelled with $v$.

□

From those two propositions, it follows that the message-based distributed consequence finding algorithm of Section 3.2 can be used to compute the maximal conjunctive rewritings of a query. This algorithm computes the set of proper prime implicates of a literal w.r.t. a distributed propositional clausal theory. Therefore, if it is applied to the distributed theory resulting from the propositional encoding of the schema of a SomeWhere network, with the extensional classes symbols as *target variables*, and triggered with a literal $\neg q$, it computes in fact the negation of the maximal conjunctive rewritings of the *atomic* query $q$. This result also holds for any arbitrary query since, in our setting, the maximal rewritings of such a query can be obtained by combining the maximal rewritings of its atomic components.

A corollary of these two propositions is that, in our setting, query answering is $BH_2$-complete w.r.t. query complexity and polynomial w.r.t. data complexity.

## 5. Experimental Analysis

This section is devoted to an experimental study of the performances of the distributed consequence finding algorithm described in Section 3, when deployed on real peer-to-peer inference systems. Particularly, the aim of our experiments is to study scalability issues of the SomeWhere infrastructure for the Semantic Web. Our experiments have been performed on networks of 1000 peers. Our goal is thus to study the practical complexity of the reasoning on networks of this size and to answer questions such as: how deep and how wide does the reasoning spread on the network of peers? Does the network cope with the traffic load? How fast are the results obtained? To what extent do the results integrate information from distinct peers? etc.

So far, large real corpus of distributed clausal theories are still missing. Since deploying new real applications at this scale requires a significant amount of time, we have chosen to perform these experiments on artificially generated instances of peer-to-peer networks. These instances are characterized by the size and the form of the local theories corresponding to each peer of the network, as well as by the size and the topology of the acquaintance graph.

Since our aim is to use the SomeWhere infrastructure for Semantic Web applications, we focus our benchmarking to instances with suitable characteristics. In particular we generate local theories supposed to encode realistic ontologies and mappings, that could be written by people to categorize their own data. Acquaintances between peers are generated in such way that the topology of the resulting graph looks realistic with respect to the acquaintances between people on the Web. Therefore we focus our generation on acquaintance graphs having the so-called *small world* property, which is admitted (Newman, 2000) as being a general property of social networks (including the Web).

In the following, we first detail in Section 5.1 the generation process of our benchmark instances, the involved parameters and how they have been allowed to vary in our experiments. In Section 5.2, a first series of experiments studies the hardness of local reasoning within a single peer by evaluating the number and the size of computed proper prime implicates. This allows us to realize the intrinsic complexity of this task and thus, of the reasoning on large scale networks of such theories. It also helps us to justify the choice of some parameter values for further experiments. Finally, Section 5.3 reports the experimental results that have been obtained concerning the scalability of Somewhere on large networks of peers.

### 5.1 Benchmark Generation

Generating satisfactory instances of peer-to-peer networks for our framework means both generating realistic propositional theories for each peer, as well as an appropriate structure for the acquaintance graph. The latter is induced by the variables shared between peer theories. In the setting of the Semantic Web, they correspond to names of atomic classes involved in mappings between peer ontologies.

#### 5.1.1 Generation of the local theories

We make the following assumptions on the ontologies and the mappings that are likely to be deployed at large scale in the future Semantic Web: the ontologies will be taxonomies of atomic classes (possibly with disjointness statements between pairs of atomic classes) ; most mappings between such ontologies are likely to state simple inclusion or equivalence between two atomic classes of two different peers, but we do not want to exclude some more complex mappings involving logical combination of classes.

As a consequence, the propositional encoding of a taxonomy results in a set of clauses of length 2. Most mappings result as well in clauses of length 2. The most complex mappings might result in longer clauses, but since any set of clauses may equivalently be rewritten as a set of clauses of length 3, we can restrict to the case where these are encoded with clauses of length 3. Clauses encoding the mappings (called *mapping clauses*) are thus only clauses of length 2 and 3, (2-clauses and 3-clauses for short). We denote by $\%3cnf$ the ratio of 3-clauses to the total number of mapping clauses. This ratio reflects in some way the degree of complexity of the mappings. In our experiments we study variations of this parameter because of its significant impact on the hardness of reasoning.

The local theories are generated in two steps. We first generate a set of $m$ random 2-clauses on a number $n$ of propositional variables and randomly select a number $t$ ($t \leq n$) of target variables (corresponding to the names of extensional classes). Mapping clauses are then generated, according to a given value $\%3cnf$ and added to the theories. Since mapping clauses induce the structure of the acquaintance graph, the way they are generated is discussed below. Peer theories are thus composed of only 2-clauses and 3-clauses. In the literature on propositional reasoning, such theories correspond to so-called $2+p$ *formulas*, where $p$ denotes the proportion of 3-clauses in the whole set of clauses (note that $p$ and $\%3cnf$ are different ratios).

#### 5.1.2 Generation of the acquaintance graph

In order to focus on realistic acquaintance graphs, we have chosen to generate random graphs with "small worlds" properties, as proposed and studied by Watts and Strogatz (1998), as well as Newman (2000). Such graphs have two properties that are encountered in social networks: first, a short path length between any pair of nodes, observed in all social relationship (for instance, the widely accepted "six-degrees of separation" between humans) and, second, a high *clustering ratio*, a measurement of the number of common neighbors shared by two neighbors (for instance, it is likely that two friends share a common subset of friends).

To generate such graphs, we first generate the pairs of peers that are connected. Following the work of Watts and Strogatz (1998), we start from a so called *k-regular ring structure* of $np$ nodes, i.e., a graph the nodes of which may be arranged as a ring, and such that each node is connected to its $k$ closest neighbors. Edges of this graph are then randomly rewired, with a given probability $pr$, by replacing one (or both) of the connected peers with another peer. It has been shown that between regular graphs ($pr = 0$) and uniform random graphs ($pr = 1$), the graphs generated with $pr = 0.1$ have "small world" properties. All acquaintance graphs used in our experiments have been generated in that way, with $pr = 0.1$. Moreover, since our aim is to evaluate the scalability of our approach, we have chosen to focus on networks of significant size. For all our instances, we have fixed the number of peers $np$ to 1000 and the number $k$ of edges per peer to 10.

Once the topology of the network has been generated, local theories of each peer are generated. Portion of the theories encoding taxonomies are first generated as previously described. Then, for each edge of the generated graph mapping clauses are added. For simplicity, we have chosen to add a fixed number $q$ of mapping clauses for each edge. Mapping clauses are randomly generated by picking one variable in each of the two peers theories and by negating them with probability 0.5. With a $\%3cnf$ probability, a third literal (randomly chosen between the two peers) is added to the clause to produce mapping clauses of length 3. As a consequence, the average number of variables shared between two connected peers is $(2 + \%3cnf) * q$.

### 5.2 Hardness of Local Reasoning within a Single Peer

In our setting, the first part of the reasoning performed at each peer consists in computing the *proper* prime implicates of the received litteral w.r.t. the local theory of the peer. In order to grasp the hardness of this local reasoning we have first conducted an experimental study to evaluate the number and the form of such implicates, and also, since our local theories are $2 + p$ clausal theories, to evaluate the impact of the ratio $p$ on these values. These experiments have been performed using a modified version of *Zres* (Simon & del Val, 2001).

Prime implicates have been already studied for 3-CNF random formulas (Schrag & Crawford, 1996). This corresponds to the case where $p = 100\%$. We first take this as a reference for comparison with *proper* prime implicates. We consider 3-CNF theories of $m = 30$ clauses over $n$ variables ($n$ ranging from 12 to 28). The left part of Figure 4 presents the characteristics of prime implicates for different values of $n$. Each curve describes how the prime implicates distribute according to their length. The curves correspond to average values over 1500 experiments. For instance, for $n = 28$ variables, there are on average more than 680 prime implicates of length 7. The right part of Figure 4 describes the size of the whole set of prime implicates by means of the cumulative distribution function of its total number $L$ of literals. Each point $(x, y)$ on a curve must be read as "over the $N = 1500$ runs, $y.N$ of them led to a $L$ value smaller than $x$". This representation is convenient for exhibiting exponential distributions. The point to stress here is that for such small formulas, the median value of the size of the whole set of prime implicates already reaches more than ten thousand literals. On some rare runs (less than 5%), the set of prime implicates has more than one hundred thousand literals. We can also observe that augmenting the number of variables increases the difficulty since it results in longer and longer prime implicates being produced in the final result. Note that a simple random 3CNF formula of only 30 clauses may already generate thousands of prime implicates. Since computing implicates for random formulas is known to require lots of subsumption checks (Simon & del Val, 2001), the final set of prime implicates may be very hard to obtain in practice.

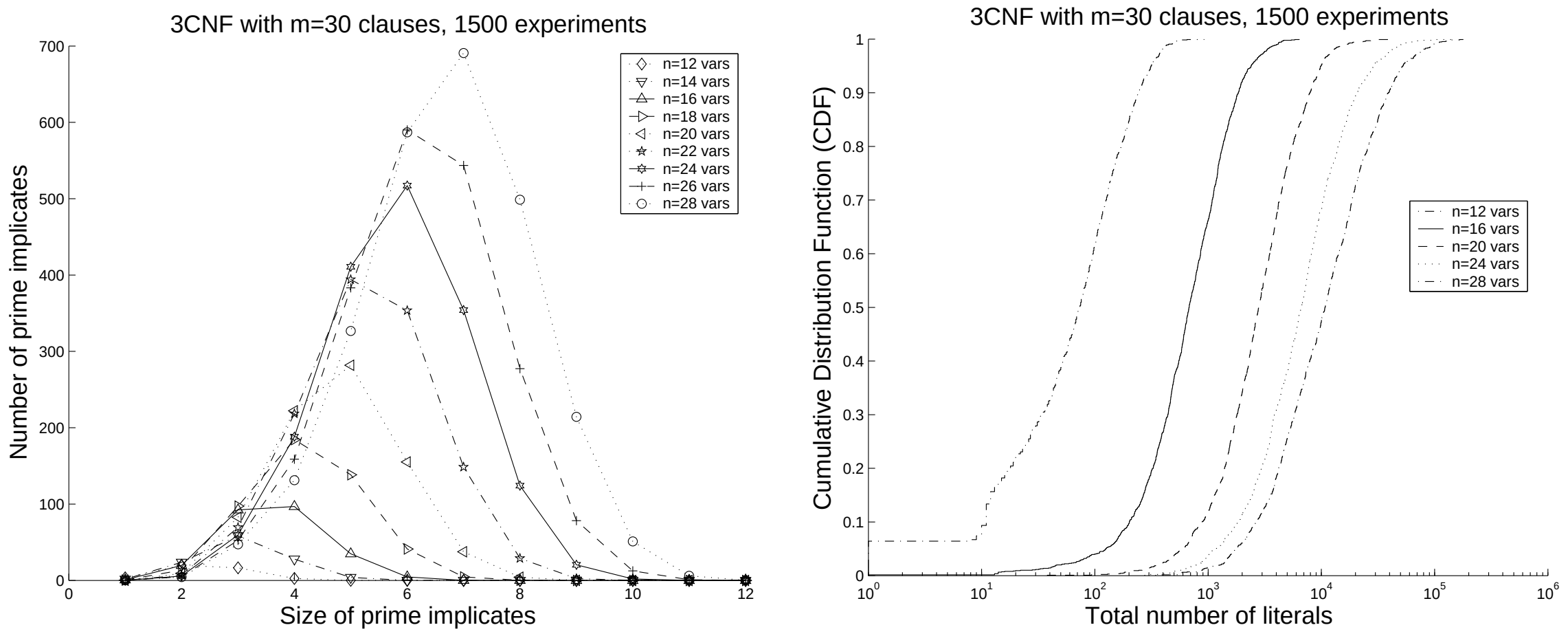

Figure 4: Prime Implicates in a uniform random 3-CNF theory

While such results are not new (Schrag & Crawford, 1996), it is interesting to compare them with those obtained in similar conditions when computing *proper* prime implicates, described on Figure 5. We can observe that curve shapes are very similar to those previously obtained but that the values are one order of magnitude smaller. Note that for $n = 28$, the median value of the size of the whole set of proper prime implicates is already about one thousand of literals. And similarly, for large values of $n$, a majority of proper prime implicates are long. Intuitively, one may explain this phenomenon by the fact that proper prime implicates are prime implicates of the initial theory augmented with an additional literal. But this literal presumably induces clauses reductions in the theory and as a consequence more subsumptions. The problem thus becomes a simplified – but not very much – version of the prime implicates problem. From these experiments, a first conclusion is that, even for very small 3-CNF theories, the number of proper prime implicates may be quite large, and some of them may be quite big.

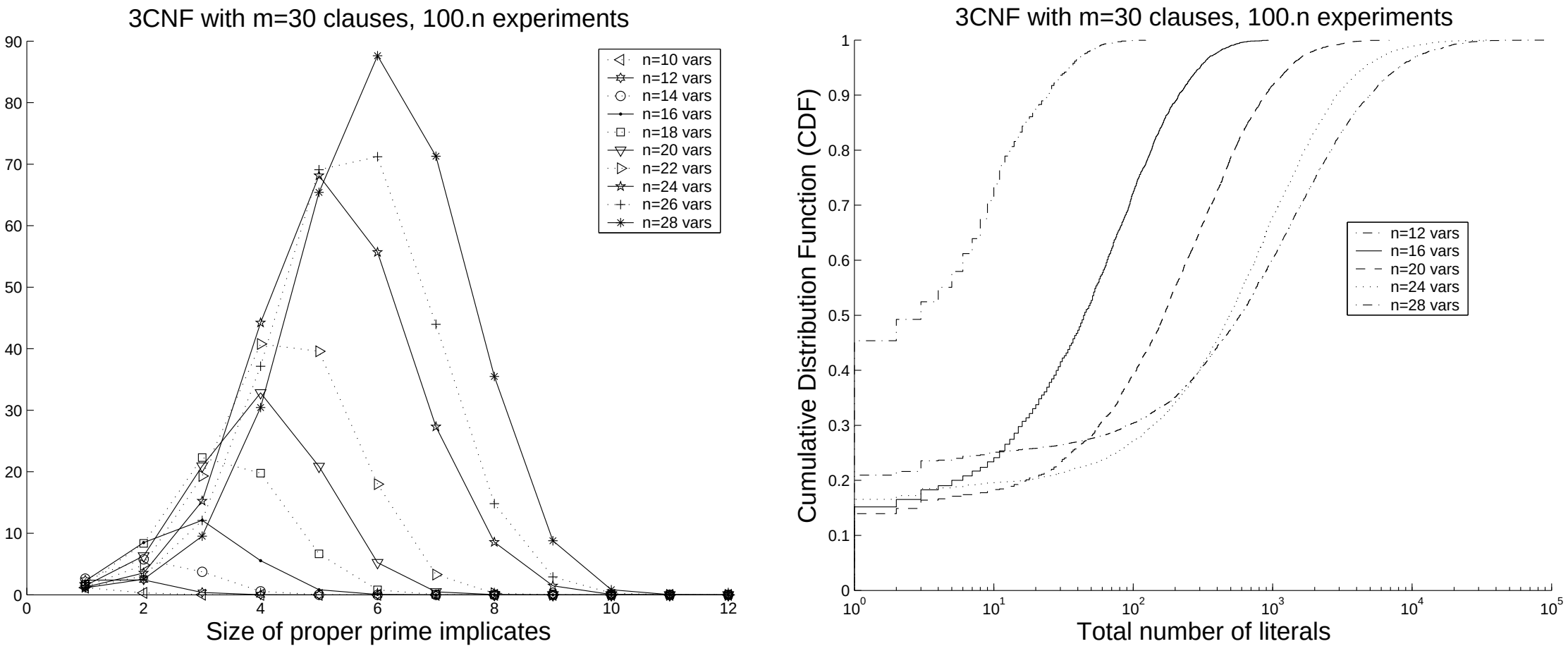

Figure 5: Proper Prime Implicates in a uniform random 3-CNF theory

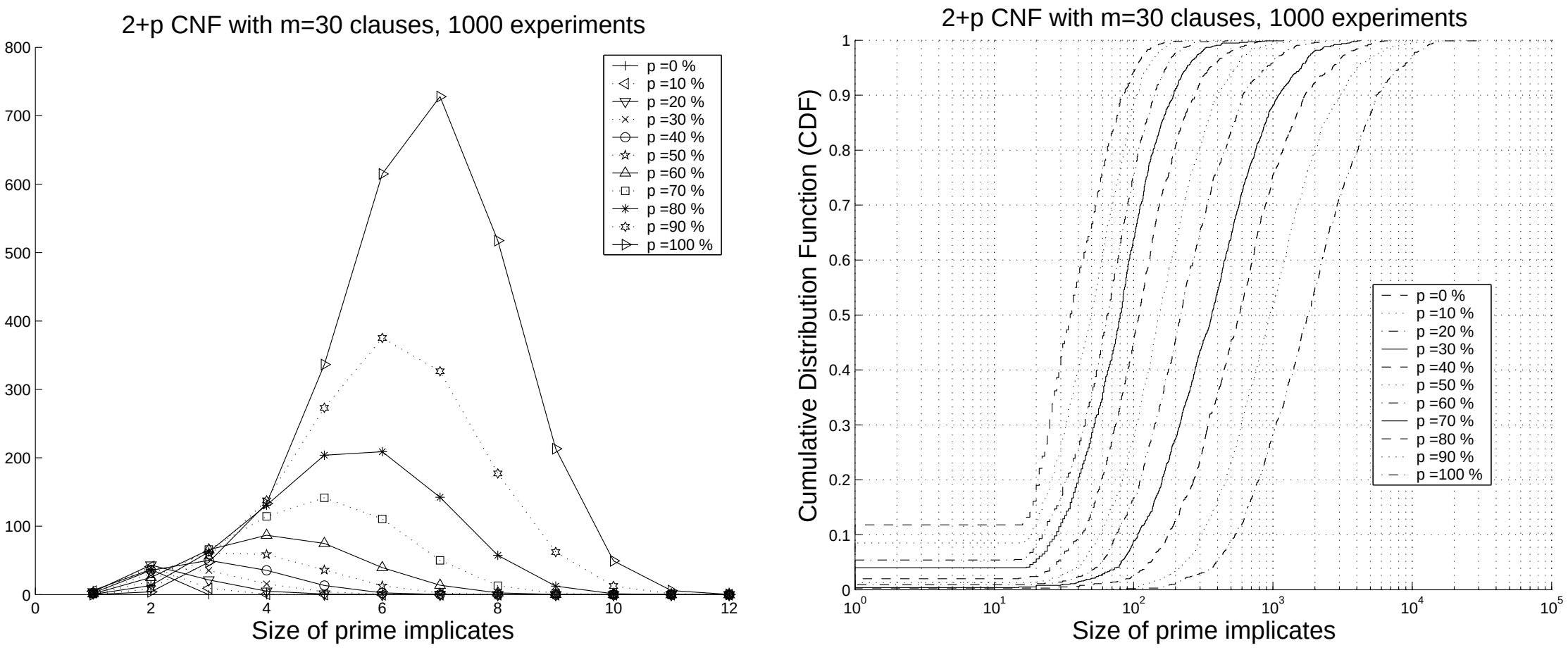

Figure 6: Prime Implicates in a uniform random $2 + p$-CNF theory ($m = 30$, $n = 28$)

Let us now focus our further experiments on local $2 + p$ CNF theories, with smaller values of $p$, supposed to better correspond to the encoding of applications in which part of the knowledge is structured as a tree or as a dag (which is encoded with 2-clauses). Figure 6 describes the prime implicates for a $2 + p$ CNF theory with $m = 30$ clauses and $n = 28$ variables, for values of $p$ ranging from 0% to 100% (the curve corresponding to the case $p = 100\%$ is the same as in Figure 4). Similarly, Figure 7 describes the characteristics of the proper prime implicates for the different values of $p$. As previously, we can observe that the curves have similar shapes. From the cumulative distribution function (CDF) curves, it appears that the hardness (measured as the total size of the prime/proper prime implicates) of the $2 + p$ CNF formula grows exponentially with the value of $p$. Even for small values of $p$ the problem may be hard. And as $p$ increases, larger and larger clauses quickly appear in the result.

Figure 8 studies the characteristics of proper prime implicates for a fixed and small value of $p = 10\%$, and increasing sizes $m$ of the theory and of the number of variables $n$.

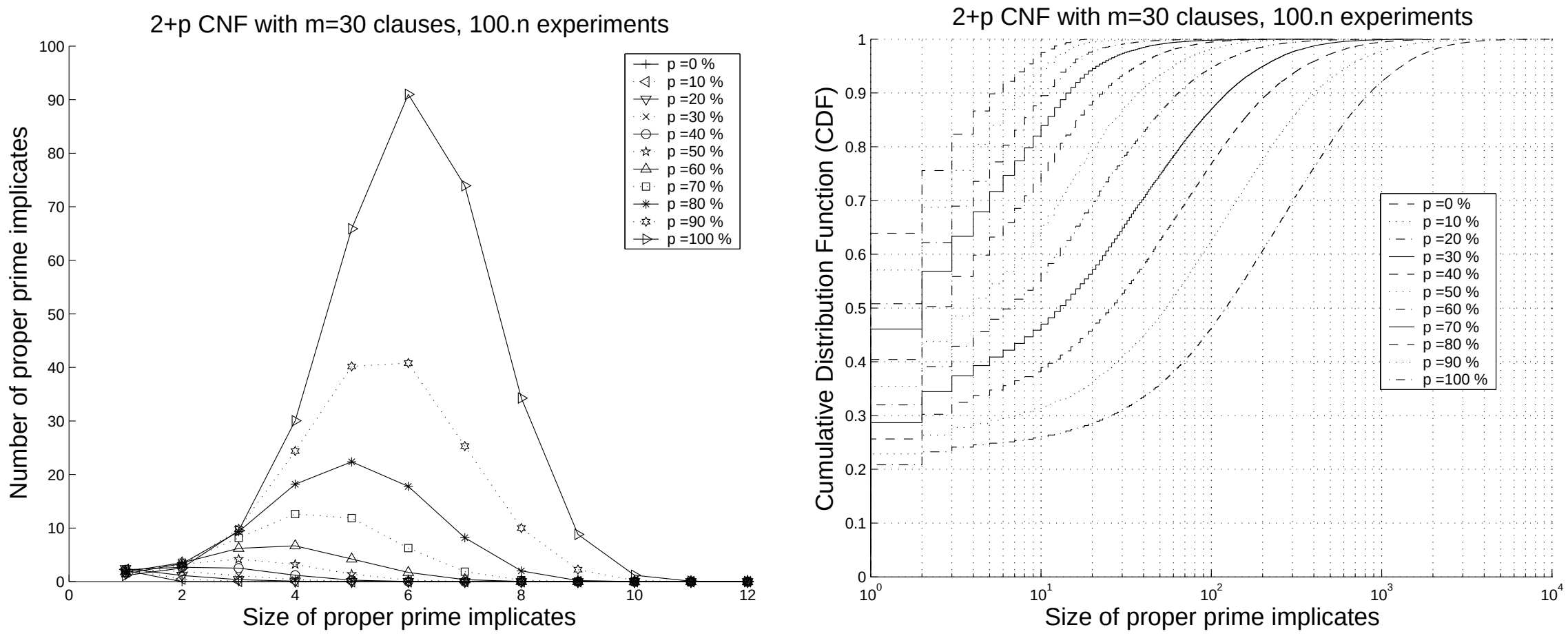

Figure 7: Proper Prime Implicates in a uniform random 2+$p$-CNF theory ($m = 30$, $n = 28$)

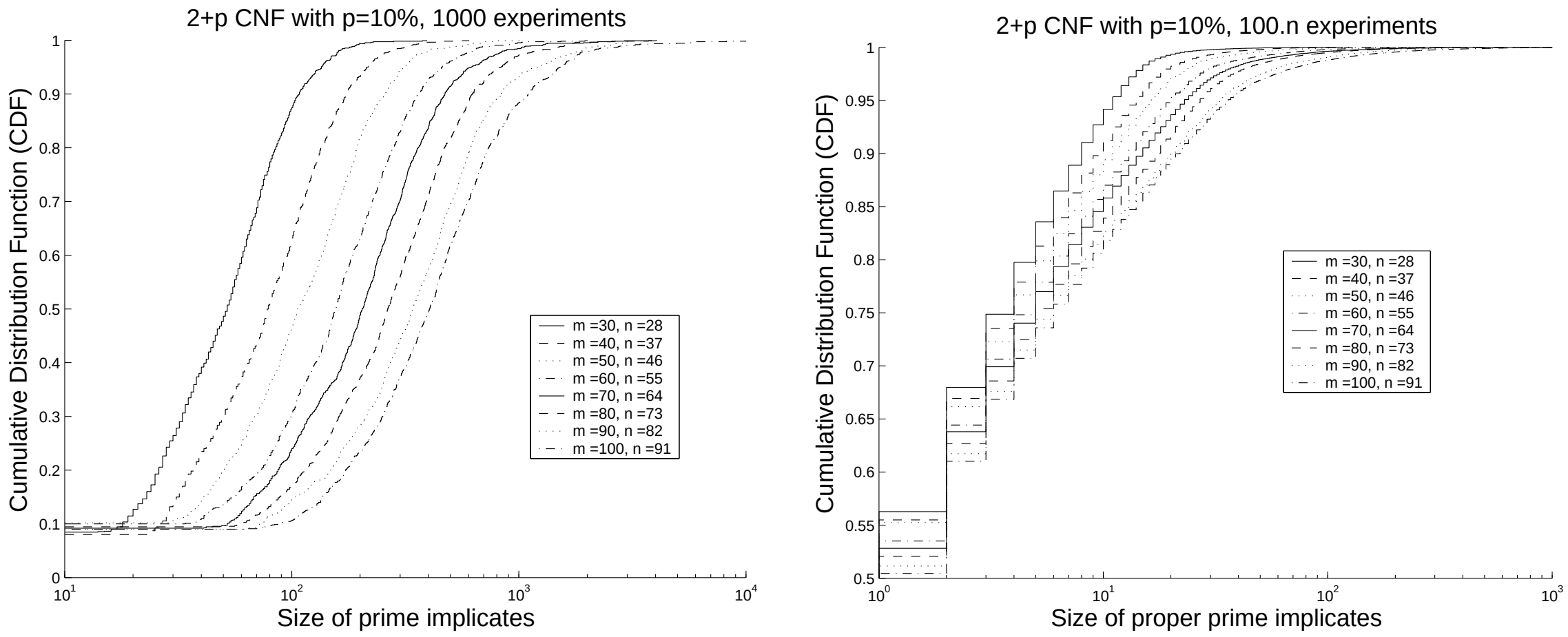

Figure 8: Size of Prime and Proper Prime Implicates in a uniform random 2+$p$-CNF theory for a fixed $p$

We have chosen a small value of $p$ to focus on the characteristics of larger theories (up to $m = 100$ clauses). It is worth noticing that even for $m = 100$ and $p = 10\%$, the problem seems simpler than for $m = 30$ and $p = 100\%$ (note that on the right part of Figure 6 the $y$-axis has been rescaled to $[0.5 - 1]$). Such kinds of peer theories seem to have a reasonable behavior from an integration perspective: half of the queries are very short and only a small part of the queries are still very hard (the exponential distribution is still observed).

### 5.3 Scalability of Distributed Reasoning

The previous section has clearly shown that the local computation performed at each peer may be really hard, even for small and simple theories. When focusing at the level of whole Somewhere networks, the splitting/recombining strategy followed by our distributed consequence finding algorithm clearly adds another source of combinatorial explosion. In order to evaluate the scalability of Somewhere networks, we have performed two kinds of experiments. The first one aims at studying the behavior of Somewhere networks during query processing. It consists in counting the number of messages circulating in a network and the number of peers that are solicited while processing a query. In particular, we have measured the distribution of the depth of query processing as well as the potential width of a query. The *depth of processing* (*depth* for short) of a query is the maximum length of the reasoning branches developed by the distributed algorithm for returning an answer to this query. The *width* of the query measures the average number of neighbors that are solicited by a peer during the reasoning. The second kind of experiments aims at evaluating the processing time and the number of answers obtained for a query.

In all our experiments, the local theories of the 1000 peers of the network have been generated as described in Section 5.1, with the fixed values $m = n = 70$ and $t = 40$. Those numbers are close to what we would obtain by encoding taxonomies having the form of balanced trees, with a depth between 3 and 5, in which each class has between 2 and 3 sub-classes, and the extensional classes of which correspond to the leaves of the tree. Each peer theories contains in addition $10 * (1 - \%3cnf) * q$ mapping clauses of length 2 and $10 * q * \%3cnf$ mapping clauses of length 3. Since we have seen in Section 5.2 that the proportion of 3-clauses in each local theory has a strong impact on the hardness of the local computing performed for each (sub)query at each peer, we have studied variations of the two related parameters: $q$ and $\%3cnf$ of increasing complexity. Note that the distributed theories considered in these experiments are quite large since the size of the corresponding centralized theories ranges from 80 000 up to 130 000 clauses over 70 000 variables.

### 5.3.1 Behavior of Distributed Query Processing

Let us now study the impact of the number $q$ of mapping clauses per edge and of the ratio $\%3cnf$ of mapping clauses of length 3, on the depth of queries. For this purpose we have measured, for each pair $(q, \%3cnf)$, the depth of 1000 random queries[1]. Since we know (cf. Section 5.2) that local computing may be sometimes very hard and therefore may require a lot of time, it has been necessary to introduce an additional *timeout* parameter. Each query is thus tagged with its remaining time to live, which, on each peer, is decreased of the local processing time, before propagating the induced subqueries. For these experiments, the timeout value has been set to 30 seconds.

Figure 9 shows the cumulative distribution functions corresponding to each pair $(q, \%3cnf)$. Each point on the figure reports a run, for a distinct query. The four leftmost curves correspond to cases where the query depth remains relatively small. For instance, for $q = 2$ and $\%3cnf = 0$ none of the 1000 queries has a depth greater than 7. Altogether on the four leftmost curves none of the queries has a depth greater than 36. This suggests that our algorithm behaves well on such networks.

When the value of $\%3cnf$ increases, queries have longer depths. For instance, with $q = 3$ and $\%3cnf = 20$, we can observe that 22% of the queries have a depth greater than 100 (the maximum being 134). The three rightmost curves have a similar shape, composed of three phases: a sharp growth, corresponding to small depth queries, followed by a plateau, and then a slower growth. The small depth query processing and the 'plateau' are characteristics of an exponential distribution of values: most of the processing is easy, but the little remaining is very hard. The slow growth observed is due to the timeout, a side-effect of which is to bound the query depth. Without such a timeout, previous experiments suggest that there would exist some queries requiring very long reasoning branches. This point is outlined on the curve corresponding to the hardest cases ($q = 3$ and $\%3cnf = 20$) where there is no query of depth between 20 and 60. This suggests that when hard processing appears, it is *very hard*. One may notice here that this last case corresponds to local theories which are very close to one of cases studied in section 5.2. As a matter of fact, with $q = 3$ and $\%3cnf = 20$ local theories are very close to a $2 + p$ theories of $m = 100$ clauses with $p = 6\%$.

[1] For convenience, and since time is not here a direct issue (except from the timeout impact), these experiments have been performed on a single computer, running the 1000 peers. This made the building of reports relying on peer traces much easier.

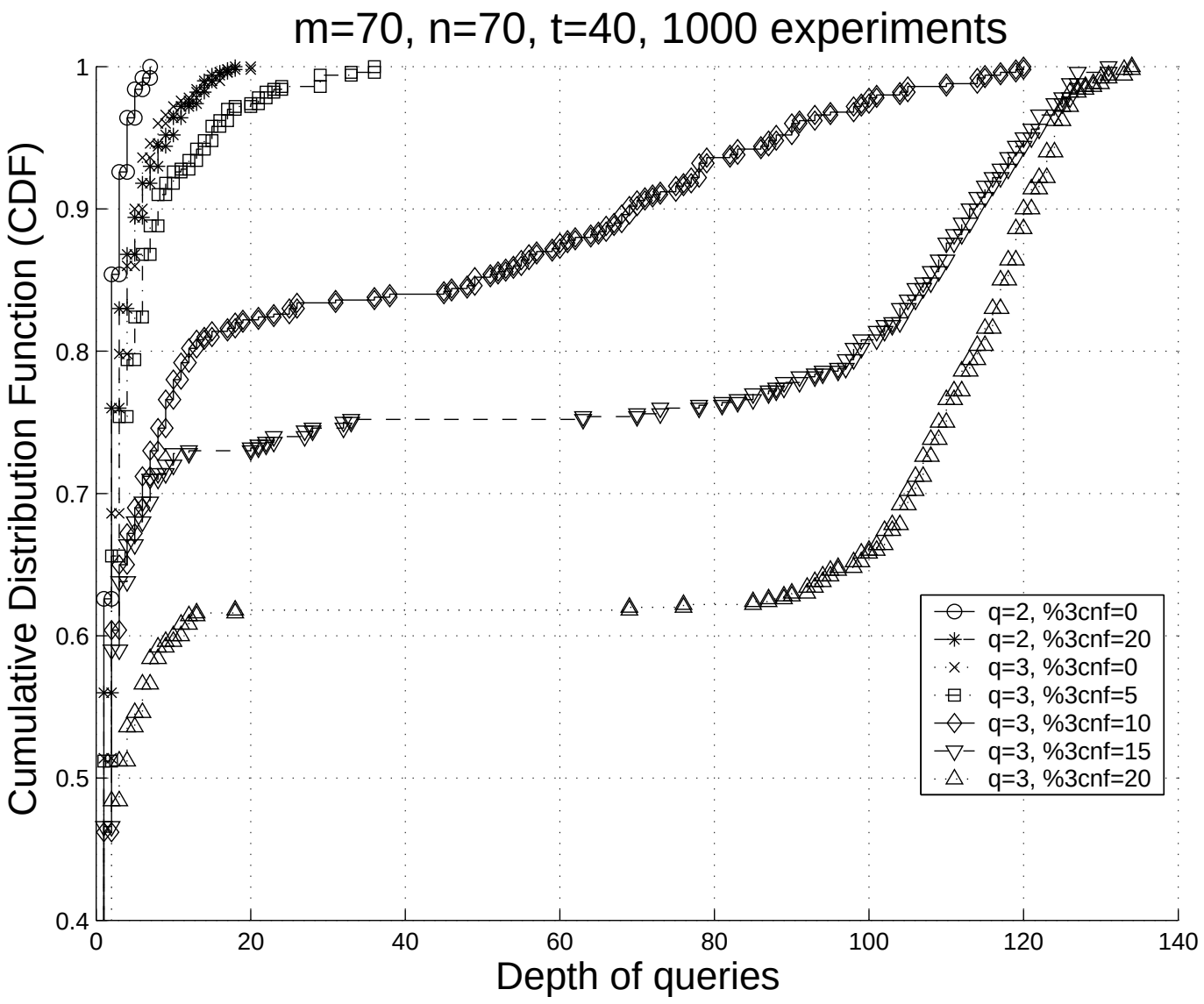

Figure 9: Cumulative distribution function of the depth of 1000 queries. $q$ is the number of mapping clauses per edge, and $\%3cnf$ is the ratio of 3-clauses in the mappings. The $y$ scale has been re-centered on $[0.4 - 1.0]$.

In other experiments that we not detail here, we have checked that such an exponential distribution of values cannot be observed when the acquaintance graphs have a strong structure of a ring (this corresponds to a rewiring probability $p = 0$ in the network generator). Because such an exponential distribution can be observed on random graphs (corresponding to $p = 1$), we suspect that such a behavior is due to the short path length between two peers, a property shared by both random and small world graphs. However, those two types of graphs differ on their clustering coefficient, a property having a direct impact on our algorithm behavior.

The depth of a query is the length of the history handled by the algorithm, in which a given peer can appear several times since the processing of subqueries (resulting of several splitting of clauses entailed by the initial query) can solicit the same peer several times.

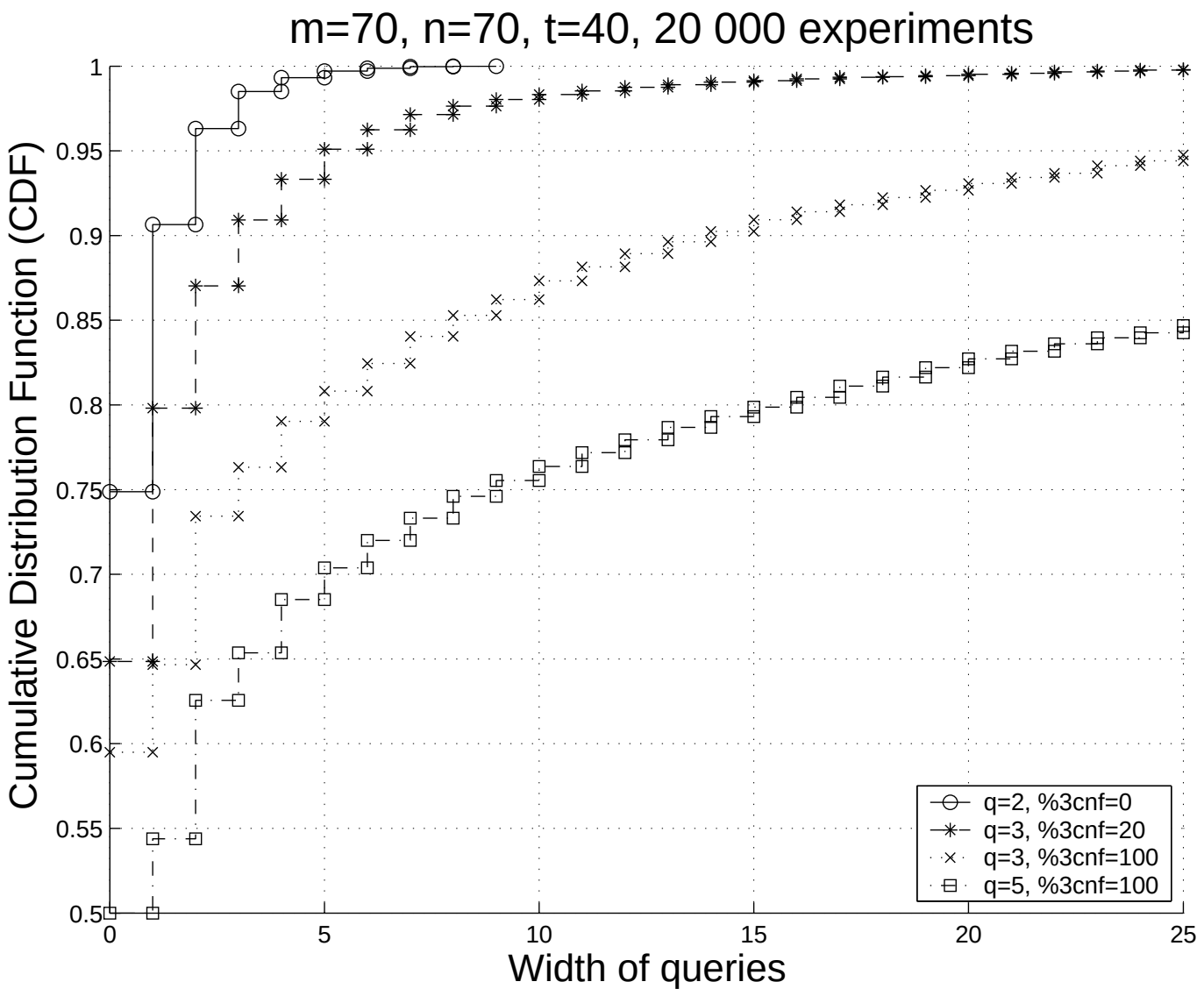

Figure 10: Cumulative Distribution Function of queries width without timeouts. Each curve summarise 20000 runs. The Y scale has been re-centered on $[0.5 - 1.0]$, the X axis on $[0 - 25]$.

We have also measured the *integration degree* of queries, which is the number of *distinct* peers involved in the query processing. We have observed the same kind of exponential distributions of values than for the depth of queries, but with 20% smaller values. This means that about one fifth of the history peers are repeated ones. That phenomenon was not observed on random acquaintance graphs and this seems closely related to the small world topology. It is important to point out that such a difference between small world and random graphs could only be observed on large experimental data, with a large number of peers.

We have also studied how wide the reasoning propagates in the network of peers during query processing. For this purpose we have evaluated the average number of neighbors peers that are solicited by a peer when solving a query. We have estimated this value by generating 20000 random queries on random peers, and counting for each of them, the number of induced subqueries to neighbor peers. Figure 10 shows, for different pairs $(q, \%3cnf)$, the corresponding cumulative distribution functions. For instance, for $q = 2$ and $\%3cnf = 0$, more than 75% of the queries are solved locally and 15% of the remaining ones are solved by asking only one neighbor. With $q = 5$ and $\%3cnf = 100$, about 25% of the queries solicit at least 10 neighbors. Of course, 25% of the subqueries may also solicit at least 10 peers and so on.

To summarize, our experiments have pointed out a direct impact of the $\%3cnf$ value on the hardness of query processing, which is not surprising considering the hardness of clauses of length 3 for prime implicates computation. Those experiments also suggest an exponential distribution of query depths, due to the short path length between two peers in the acquaintance graphs, and with an important repetition of solicited peers, due to the large clustering coefficient of small world acquaintance graphs.

### 5.3.2 Time and Number of Answers

We now report a time performance study of our algorithm when it is deployed on a real cluster of 75 heterogeneous computers[2]. Based on the observations of the previous sections, we have chosen to focus on 5 differents kinds of acquaintance graphs, denoted *Very Easy*, *Easy*, *Medium*, *Hard* and *Very Hard* (see Table 1). One of the main goals of this section is to estimate where the limits of processing of our algorithm are when it faces with hard (and even very hard) Somewhere networks. Again, for all these experiments, we have set the timeout value to 30s.

| Network | Very Easy $q = 2$ $\%3cnf = 0$ | Easy $q = 3$ $\%3cnf = 20$ | Medium $q = 3$ $\%3cnf = 100$ | Hard $q = 5$ $\%3cnf = 100$ | Very Hard $q = 10$ $\%3cnf = 100$ |
|---|---|---|---|---|---|
| $1^{st} ans.$ | 0.04s (100%) | 1.26s (99.6%) | 1.58s (95.6%) | 1.39s (89.3%) | 2.66s (49.7%) |
| $10^{th} ans.$ | 0.06s (14.3%) | 1.37s (25.6%) | 0.99s (33.3%) | 1.13s (12.0%) | 5.38s (29.9%) |
| $100^{th} ans.$ | – | 2.11s (12.7%) | 0.84s (27.0%) | 4.09s (10.7%) | 11.0s (9.0%) |
| $1000^{th} ans.$ | – | 4.17s (6.80%) | 4.59s (21.2%) | 11.35s (7.15%) | 16.6s (1.80%) |
| all | 0.07s | 5.56s | 14.6s | 21.23s | 27.74s |
| % timeout | 0% | 13.9% | 37.5% | 66.9% | 86.9% |
| #answers | 5.17 | 364 | 1006 | 1004 | 65 |
| %unsat | 4.4% | 3.64% | 3.76% | 1.84% | 1.81% |

Table 1: Characteristics of the query processing ranging from very easy to very hard cases.

The values reported in Table 1 are mean values over more than three hundred different random queries. Each column indicates the time needed to produce respectively the $1^{st}$, $10^{th}$, $100^{th}$ and $1000^{th}$ answer of a query. The mean time (in seconds) is followed by the percentage of initial queries that are taken into account in the average. For instance, for a medium case with $q = 3$, 12.7% of the queries have produced more than 100 answers, and the $100^{th}$ answer was given on average after 2.11 seconds (the average does not take into account queries that did not produce at least 100 answers). The *all* row corresponds to the mean time needed to produce all answers, including queries that lead to timeout, the percentage of which is reported in the *%timeout* row. The last two rows report the mean number of answers and the ratio of proved unsatisfiable queries w.r.t. the Somewhere network (some unsatisfiable queries w.r.t. the network may not have been counted since inconsistency might have been found after the timeout).

Unsurprisingly, no timeout occurs for the *Very Easy* case. It is known that satisfiability checking and prime implicates computation are tractable for sets of clauses of length 2. Moreover, the high partitioning of the global theory induced by the low value of $q$ (number of mappings per peer) is often a witness of "easy" cases for reasoning for centralized theories.

[2] All computers were Linux Athlon units, with 1GB of RAM. 26 of them ran at 1.4GHz, 9 at 1.8GHz and 40 at 2 GHz. Each computer was running around 14 randomly assigned peers

The point to outline for this case is that there are 5 answers on average, and that they are produced very quickly by our algorithm (in less than 0.1 seconds).

Let us now point out that, even on *Medium* and *Hard* instances, our algorithm produces a lot of answers. For instance, we have obtained an average of 1006 answers for $q = 3$, and of 1004 answers for $q = 5$. In addition, on those hard instances, 90% of runs produced at least one answer. It is noticeable that in the *Very Hard case* ($q = 10$), half of the queries produce at least one answer, even if only 13% of them do complete without a timeout. Let us note yet that checking the satisfiability of the corresponding centralized theories can also be very hard. As a matter of fact the formula corresponding to the centralized version of all the distributed theories has $n$=70 000 variables and $m$=120 000 clauses, 50 000 of which are of length 3. The ratio of 3-clauses in those $2 + p$ theories is thus $p = 0.416$. It has been shown (Monasson, Zecchina, Kirkpatrick, Selman, & Troyansky, 1999) that, for $2 + p$ random formulas, if one does not restrict the locality of variables, the SAT/UNSAT transition is continuous for $p < p_0$ ($p_0 = 0.41$) and discontinuous for $p > p_0$, like in 3-SAT instances. Intuitively, for $p > p_0$, the random $2 + p$-SAT problem shares the characteristics of the random 3-SAT problems, which is well known as the canonical NP-Complete problem. Let us recall that our generation model induces a high clustering of variables inside each peer. Therefore we cannot claim that the corresponding centralized theories have exactly the same cararacteristics as uniform $2+p$-SAT random formulas. However, if one only focus on the values of the parameters $m$ and $n$, for the characteristics of our *Very Hard* network, the transition phase between SAT and UNSAT instance occurs at $m/n$=1.69. Here, we have $m/n$=1.71, which is close enough from the transition phase to suspect that this is where hard instances may be found.

To summarize, when deployed on a real cluster of heterogeneous computers, our algorithm scales very well. Even on *Very Hard* instances that share some characteristics of a very large $2 + p$-SAT formula at the crossover between the 2-SAT/3-SAT and the SAT/UNSAT transitions, our algorithm is able to return many answers in a reasonable time.

## 6. Related work

In Section 6.1, we situate our work w.r.t. existing work related to distributed reasoning or distributed logics, while in Section 6.2 we summarize the distinguishing points of SomeWhere among the existing peer data management systems.

### 6.1 Related Work on Distributed Reasoning

The message passing distributed algorithm that we have described in Section 3 proceeds by splitting clauses and distributing the work corresponding to each piece of clause to appropriate neighbor peers in the network. The idea of splitting formulas into several parts may be found back as the so called "splitting rule" in the natural deduction calculus, introduced in the middle of the 1930's by two independent works of Gentzen (1935, 1969) and Jaśkowski (1934). Our algorithm may be viewed as a distributed version of Ordered Linear Deduction (Chang & Lee, 1973) to produce new target clauses. This principle has been extended by Siegel (1987) in order to produce all implicates of a given clause belonging to some target language, and further extended to the first order case by Inoue (1992).

We have already pointed out the differences between our work and the approach of Amir and McIlraith (2000). In a peer-to-peer setting, tree decomposition of the acquaintance graph is not possible. In addition, as we have shown in the introductory example, the algorithm of Amir and McIlraith (2000) is not complete in the general case for proper prime implicate computation. However, Goasdoué and Rousset (2003) have shown that completeness can be guaranteed for a family of P2PIS with peer/super-peers topology. It describes how to encode a P2PIS with peer/super-peers into a partitioned propositional theory in order to use the consequence finding algorithm of Amir and McIlraith (2000). The model-based diagnosis framework for distributed embedded systems (Provan, 2002) is based on the work of Amir and McIlraith (2000). We think it can benefit from our approach to apply to a real peer-to-peer setting in which no global knowledge has to be shared.

Other forms of distributed reasoning procedures may be found in multiagent frameworks, where several agents try to cooperate to solve complex problems. Problems addressed in this way can generally be decomposed in several interacting subproblems, each of which is addressed by one agent. This is the case for Distributed Constraint Satisfaction Problems (DCSP, Yokoo, Durfee, Ishida, & Kuwabara, 1998). Given a set of variables, each of them being associated to a given domain, and a set of constraints on theses variables, the problem is to assign each variable one value in its respective domain, in such a way that all constraints are satisfied. In the distributed case, each variable is associated to some agent. Constraints may either concern a set of variables relevant to a same agent or variables relevant to different agents. In the first case, they may be considered as characterizing the local theory of the agent, while in the second case they may be assimilated to mapping constraints between agents. Each mapping constraint is assigned to one agent. The problem addressed in DCSP is easier than the problem of consequence finding since it is a satisfiability problem, which is NP-complete. While centralized CSP are solved using a combination of backtrack search and consistency techniques, algorithms used to solve DCSP use asynchronous versions of backtracking (Yokoo, Durfee, Ishida, & Kuwabara, 1992; Yokoo et al., 1998) and consistency techniques (Silaghi, Sam-Haroud, & Faltings, 2000). Basically, agents proceed by exchanging invalid partial affectations, until converging to a globally consistent solution. Similar ideas may also be found in distributed ATMS (Mason & Johnson, 1989), where agents exchange nogood sets in order to converge to a globally consistent set of justifications. Let us note that in contrast with the peer-to-peer vision, such methods aim at sharing some global knowledge among all agents.

Probabilistic reasoning on bayesian networks (Pearl, 1988) has also given rise several adaptations suited to distributed reasoning (e.g., the message passing algorithm of Pfeffer & Tai, 2005). However the problem addressed in this context is different, since it consists in updating a set of a posteriori beliefs, according to observed evidence and a set of conditional probabilities. These conditional probabilities are of the form $P(x|u_1, ..., u_n)$ and describe the probability of the event $x$ when $u_1$ and ... and $u_n$ are observed. They describe value interactions that can be viewed as mappings between a conjunction of literals and a single literal, which are oriented because of the nature or conditioning.

Another work using oriented mappings is the framework of *distributed first order logic* (Ghidini & Serafini, 2000), in which a collection of first order theories can communicate through *bridge rules*. This approach adopts an epistemic semantics where connections between peer theories are reflected by mappings between the respective domains of inter-pretation of the involved peers. Based on that work, *distributed description logics* have been introduced by Borgida and Serafini (2003) and a distributed tableau method is described for reasoning in distributed description logics has been proposed by Serafini and Tamilin (2004b). A reasoning algorithm has been implemented in Drago (Serafini & Tamilin, 2004a) where the bridge rules are restricted to inclusion statements between atomic concepts.

### 6.2 Related Work on Peer Data Management Systems

As we have pointed it out in Section 4, the Somewhere peer data management system distinguishes from Edutella (Nejdl et al., 2002) by the fact that there is no need of super-peers, and from Piazza (Halevy et al., 2003b, 2003a) because it does not require a central server having the global view of the overlay network.

From the semantic point of view, Somewhereuses a propositional language and mappings correspond to unrestricted formulas. Its semantics is the standard propositional semantics and query answering is always decidable. The peer data management systems investigated by Halevy et al. (2003b) use a relational language and their mappings correspond to inclusion statements between conjonctive queries. The semantics used in this work is the standard FOL semantics, for which query answering is shown to be undecidable in the general case. Restricting to acyclic mappings renders query answering decidable in Piazza, but checking this property requires some global knowledge on the network topology and is performed by the central server.

The peer data management system considered in the work of Calvanese et al. (2004) is similar to that of Halevy et al. (2003b) but proposes an alternative semantics based on epistemic logic. With that semantics it is shown that query answering is always decidable (even with cyclic mappings). Answers obtained according to this semantics correspond to a subset of those that would be obtained according to the standard FOL semantics. However, to the best of our knowledge, these results are not implemented.

From the systems point of view, the recent work around the coDB peer data management system (Franconi, Kuper, Lopatenko, & Zaihrayeu, 2004) supports dynamic networks but the first step of the distributed algorithm is to let each node know the network topology. In contrast, in Somewhere no node does have to know the topology of the network.

The KadoP system (Abiteboul, Manolescu, & Preda, 2004) is an infastructure based on distributed hash tables for constructing and querying peer-to-peer warehouses of XML resources semantically enriched by taxonomies and mappings. The mappings that are considered are simple inclusion statements between atomic classes. Compared to KadoP (and also to Drago, Serafini & Tamilin, 2004a), the mapping language that is dealt with in Somewhere is more expressive than simple inclusion statements between atomic classes. It is an important difference which makes Somewhere able to *combine* elements of answers coming from different sources for answering a query, which KadoP or Drago cannot do.

Somewhere implements in a simpler setting the (not implemented) vision of peer to peer data management systems proposed by Bernstein et al. (2002) for the relational model.

## 7. Conclusion

The contributions of this paper are both theoretical and practical. We have provided the first distributed consequence finding algorithm in a peer-to-peer setting, and we have exhibited a sufficient condition for its completeness. We have developed a P2PIS architecture that implements this algorithm and for which the first experimental results look promising. This architecture is used in a joint project with France Télécom, which aims at enriching peer-to-peer web applications with reasoning services (e.g., Someone, Plu et al., 2003).

So far, we have restricted our algorithm to deal with a vocabulary-based target language. However, it can be adapted to more sophisticated target languages (e.g., implicates of a given, maximal, length, language based on literals and not only variables). This can be done by adding a simple tag over all messages to encode the desired target language.

Another possible extension of our algorithm is to allow more compact representation of implicates, as proposed by Simon and del Val (2001). This work relies on an efficient clause-distribution operator. It can be adapted by extending messages in our algorithm in order to send *compressed* sets of clauses instead of one clause as it is the case right now, without changing the deep architecture of our algorithm.

In the Semantic Web direction, we plan to deal with distributed RDF(S) resources shared at large scale. RDF(S) (Antoniou & van Harmelen, 2004) is a W3C standard for annotating web resources, which we think can be encoded in our propositonal setting.

In the distributed reasoning direction, we plan to consider more sophisticated reasoning in order to deal with a real multi-agent setting, in which possible inconsistencies between agents must be handled.

## Acknowledgnents

This paper is a revised and extended version of an ECAI short paper (Adjiman, Chatalic, Goasdoué, Rousset, & Simon, 2004) and of an IJCAI paper (Adjiman, Chatalic, Goasdoué, Rousset, & Simon, 2005).